\def\year{2021}
\documentclass[letterpaper]{article} 
\usepackage{aaai21}  
\usepackage{times}  
\usepackage{helvet} 
\usepackage{courier}  
\usepackage[hyphens]{url}  
\usepackage{graphicx} 
\usepackage{natbib}  
\usepackage{caption} 
\usepackage{hyperref}

\usepackage{url}            
\usepackage{amsmath,amssymb} 
\usepackage{multirow}
\usepackage{wrapfig}
\usepackage{multicol}
\usepackage[title]{appendix}
\usepackage[table,xcdraw]{xcolor}

\def\figvspace{\vspace{0mm}}

\def\fig#1{Figure~\ref{fig:#1}}
\def\tab#1{Table~\ref{tab:#1}}

\def\Eq#1{Eq.~(\ref{eq:#1})}

\def\mypar#1{\vspace{0mm}{\noindent\bf #1.}\hspace{1mm}}

\title{Anytime Inference with Distilled Hierarchical Neural Ensembles}
\author {
   Adria Ruiz\textsuperscript{\rm 1},
    Jakob Verbeek\textsuperscript{\rm 2} \\
}
\affiliations {
    \textsuperscript{\rm 1} Institut de Robòtica i Informàtica Industrial, CSIC-UPC
    \\
   \textsuperscript{\rm 2} Facebook AI Research\\
   aruiz@iri.upc.edu
}

\begin{document}

\maketitle

\begin{abstract}
Inference in deep neural networks can be  computationally expensive, and networks capable of anytime inference are  important in  
scenarios where the amount of compute or quantity of input data varies over time. 
In such networks the inference process can interrupted to provide a result faster, or continued to obtain a more accurate result. 
We propose Hierarchical Neural Ensembles (HNE), a novel framework to embed an ensemble of multiple networks in a hierarchical tree structure, sharing intermediate layers. 
In HNE we control the complexity of inference on-the-fly by evaluating more or less models in the ensemble.
Our second contribution is a novel hierarchical distillation method to boost the prediction accuracy of small ensembles.
This approach leverages the nested structure of our ensembles, to optimally allocate accuracy and diversity across the individual models. 
Our experiments show that, compared to previous anytime inference models, 
HNE provides state-of-the-art accuracy-computate trade-offs on the  CIFAR-10/100 and ImageNet datasets. 
\end{abstract}

\section{Introduction}
Deep learning models typically require a large amount of computation  during inference, limiting their deployment in edge devices such as mobile phones or autonomous vehicles. 
For this reason,  methods based on network pruning \cite{huang2018condensenet}, architecture search \cite{tan2019mnasnet}, as well as  manual network design \cite{sandler2018mobilenetv2}, have all been used to find more efficient model architectures. 
Despite the promising results achieved by these approaches, there exist several applications where, instead of deploying a single efficient network, we are interested in dynamically adapting the inference latency depending on external constraints.
Examples include scanning of incoming data streams on online platforms or autonomous driving, where either the amount of data to be processed or the number of concurrent processes is non-constant. In these situations, models must be able to scale the number of the operations on-the-fly depending on the amount of available compute at any point in time.
In particular, we focus on methods capable of anytime inference, \textit{i.e.} methods where the inference process can be interrupted for  early results, or continued for more accurate results \cite{huang2017multi}. 
This contrasts with other methods, where the accuracy-speed trade-off has be decided before the computation for inference starts.

\begin{figure}[t]
\begin{center}
\includegraphics[width=0.95\linewidth]{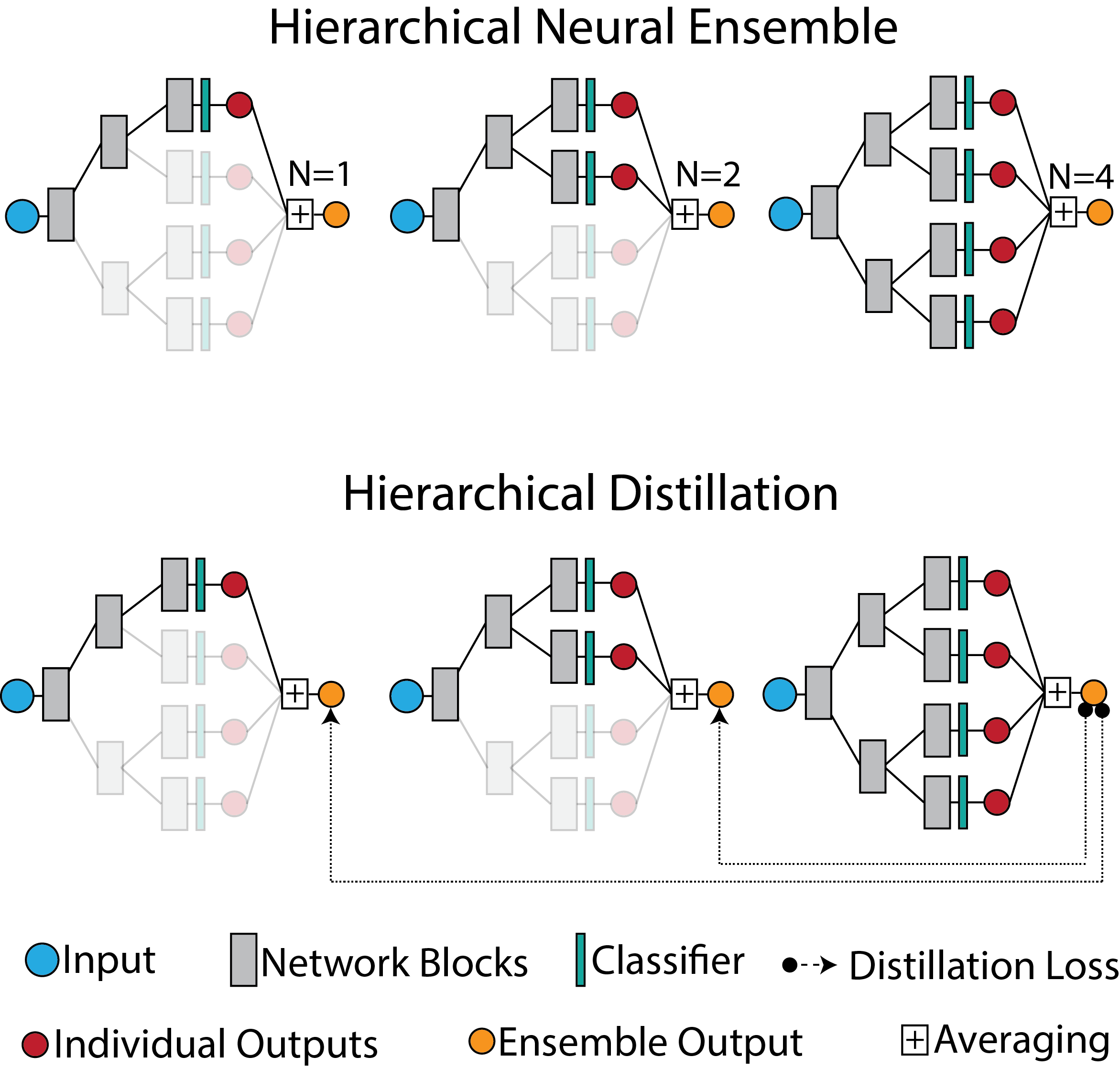}
\end{center}
\caption{
(Top) HNE shares parameters and computation in a hierarchical 
manner.  
Tree leafs represent  separate models in the ensemble. 
Anytime inference is obtained via  depth-first traversal of the tree, and using at any given time the ensemble prediction of the $N$  models evaluated so far. (Bottom) Hierarchical distillation 
leverages the full ensemble to supervise parts of the tree that are used in small ensembles. }
\label{fig:intro_figure}
\figvspace
\end{figure}

We address this problem by introducing 
Hierarchical Neural Ensembles (HNE). 
Inspired by ensemble learning \cite{breiman1996bagging}, HNE embeds a large number of networks whose combined outputs provide a more accurate prediction than any individual model. 
To reduce the computational cost of evaluating the networks, HNE employs a binary-tree structure to share a subset of intermediate layers between the different models. 
This scheme allows to control the inference complexity by deciding  how many networks to use, \textit{i.e.} how many branches of the tree to evaluate. 
To train HNE, we  propose a novel distillation method adapted to its hierarchical structure.
See \fig{intro_figure} for an overview of our approach.  

Our  contributions are summarised as follows: 
(i) To the best of our knowledge, we are the first to explore hierarchical ensembles for deep models with any-time prediction. 
(ii) We propose a hierarchical distillation scheme to increase the accuracy of ensembles for adaptive inference cost. 
(iii) 
Focusing on image classification, we show that our framework can be used to design efficient CNN ensembles. 
In particular, we evaluate the different proposed components by conducting  ablation studies on CIFAR-10/100 datasets. 
Compared to previous anytime inference methods, HNE  provides state-of-the-art accuracy-speed trade-offs on the CIFAR  datasets as well as the more challenging ImageNet dataset.

\section{Related Work}

\mypar{Efficient networks}
Different approaches have been explored to reduce the inference complexity of deep neural networks.
These include the  design of efficient convolutional blocks \cite{howard2017mobilenets,ma2018shufflenet,sandler2018mobilenetv2},  neural architecture search (NAS) \cite{tan2019mnasnet,wu2019fbnet,cai2020once,yu2020bignas} and network pruning techniques~\cite{huang2018condensenet,liu2017learning}. In order to adapt the inference cost, other methods have proposed different mechanisms to reduce the number of feature channels \cite{yu2019universally,yu2018slimmable} or skip intermediate layers in a data-dependent manner~\cite{veit2018convolutional,wang2018skipnet,wu2018blockdrop}. Whereas these approaches are effective to reduce the resources required by a single network, the desired speed-accuracy trade-off need to be selected before the inference process begins.

\mypar{Anytime inference} In order to provide outputs at early inference stages, previous  methods have considered to introduce intermediate classifiers on hidden network layers ~\cite{bolukbasi2017adaptive,elbayad20iclr,huang2017multi,li2019improved,zhang2018graph}. In particular, \cite{huang2017multi} proposed a Multi Scale DenseNet architecture (MSDNet) where early-exit classifiers are used to compute predictions at any point during evaluation. More recently, MSDNets have been extended by using improved training techniques \cite{li2019improved} or exploiting multi-resolution inputs \cite{yang2020resolution}.  
On the other hand, Convolutional Neural Mixtures \cite{ruiz2019adaptative} proposed a densely connected network that can be dynamically pruned.
Finally, early-exits have been also combined with NAS by \citet{zhang2018graph} to automatically find the optimal position of the classifiers.  

Different from previous works relying on early-exit classifiers, we address anytime inference by exploiting hierarchical network ensembles. 
Additionally, our framework can be used with any base model in contrast to previous approaches which require specific network architectures \cite{huang2017multi,ruiz2019adaptative}. Therefore, our method is complementary to approaches relying on manual design, neural architecture search, or network pruning.

\mypar{Network ensembles} 
Ensemble learning  is a classic approach to improve generalization~\cite{hansen90pami,rokach2010ensemble}. 
The success of this strategy relies on the reduction in variance resulting from averaging the output of different learned predictors \cite{breiman1996bagging}. Seminal works \cite{hansen90pami,krogh1995neural,naftaly1997optimal,zhou2002ensembling} 
observed that 
a significant accuracy boost could be achieved by averaging the outputs of independently trained networks. Recent deep CNNs have also been shown to benefit from this strategy~\cite{geiger20jsm,ilg2018uncertainty,lan2018knowledge,lee2019robust,lee2015m,malinin2019ensemble}. 
A main limitation of deep network ensembles, however, is the  linear increase in training and inference costs with the number of models in the ensemble. 
Whereas some strategies have been proposed to decrease the training time~\cite{huang2017snapshot,loshchilov2016sgdr}, the high inference cost still remains as a bottleneck in scenarios where computational resources are limited. 
In this context, different works~\cite{lan2018knowledge,lee2015m,minetto2019hydra} have proposed to build ensembles where the individual networks share a subset of parameters  in order to reduce the inference cost. 

Building on these ideas, our HNE uses a binary-tree structure to share intermediate layers between individual networks. Whereas hierarchical structures have been explored for different purposes, such as learning expert mixtures \cite{liu2019deep,tanno2018adaptive,kim2017splitnet}, incremental learning \cite{roy2020tree}, or
ensemble implementation \cite{lee2015m,zhang2018deep}, our work is the first to leverage this structure for anytime inference.

\mypar{Diversity in network ensembles} 
Ensemble performance is affected by two factors
\cite{ueda1996generalization}: the accuracy of individual models, and the variance among the  model predictions. 
Different works 
encourage model diversity  by sub-sampling different portions of training data during optimization \cite{lakshminarayanan2017simple,lee2015m} or using regularization mechanisms \cite{chen2009regularized}. Using these strategies, however, the performance of each individual model is significantly reduced \cite{lee2015m}. 
For this reason, we  instead use a simple but effective strategy to encourage model diversity. 
In particular, we train our HNE by using a different initialization for the parameters of each parallel branch in the tree structure. 
Previous work~\cite{huang2017snapshot,neal2018modern} has shown that networks trained from different initializations exhibit a significant variance in their predictions.

\mypar{Knowledge distillation} 
The accuracy of a low-capacity ``student'' network can be improved by   training it on soft-labels  generated by a high-capacity ``teacher'' network, rather than directly on the training data~\cite{ba2014deep,hinton2015distilling,romero2014fitnets}. 
Knowledge distillation from the activation of intermediate network layers  \cite{sun19emnlp}, and  from soft-labels  provided by one or more networks with the same architecture as the student ~\cite{furlanello2018born} 
have also been shown to be effective.
In co-distillation~\cite{lan2018knowledge,zhang2018deep,bhardwaj2019memory}  the distinction between student and teacher networks is lost and, instead, models are jointly optimized and distilled online. For network ensembles, co-distillation has been shown effective to improve the accuracy of the individual models by transferring the knowledge from the full ensemble \cite{anil2018large,song2018collaborative}.

To improve the performance of HNE, we introduce hierarchical distillation.  Different from existing  co-distillation strategies \cite{lan2018knowledge,song2018collaborative}, our approach transfers the knowledge from the full model to smaller sub-ensembles in a hierarchical manner. 
Our approach is specifically designed for neural ensembles, where the goal is not only to improve predictions requiring a low inference cost, but also to preserve the diversity between the individual network outputs.

\section{Hierarchical Neural Ensembles}

HNE embed an ensemble of deep networks computing an output $\mathbf{y}^E = \frac{1}{N} \sum_{n=1}^{N} F_n (\mathbf{x}; {\theta_n})$ from an input $\mathbf{x}$, where $F_n$ is a network with parameters $\theta_n$ and $N$ is the total number of models. Furthermore, we assume that each network is a composition of $B+1$ functions, or ``blocks'' as $F_n (\mathbf{x}; {\theta_n}) = f_{\theta^B_n} \circ \dots \circ f_{\theta^1_n} \circ f_{\theta^0_n}(\mathbf{x})$, where each block $f_{\theta^b_n} (\cdot)$ is a set of layers with parameters ${\theta^b_n}$. 
Typically, $f_{\theta^b_n}(\cdot)$ contains operations such as convolutions, batch normalization layers, and activation functions. 

\mypar{Hierarchical sharing of parameters and computation} 
If we use different  parameters $\theta_n^b$  for all blocks and networks, then the  set of the ensemble parameters is given by ${\Theta}=\{\theta^0_{1:N},{\theta}^1_{1:N},  \dots, \theta^B_{1:N}\}$, and the inference cost of computing the ensemble output is  equivalent to evaluate $N$ independent networks. 
In order to reduce the computational complexity, 
we design HNE to share parameters and  computation employing a binary tree structure, where each node of the tree represents a computational block.
Each of the $N=2^B$ paths from the root to a leaf represents a different model composed of $B+1$ computational blocks.
The first (root) computational block is shared among all models, and after each block the computational path is continued along two branches, each with a different set of parameters from  the next   block onward.
See \fig{intro_figure} for an illustration.
Therefore, for each block $b$ there are $K=2^b$ independent sets of parameters $\theta^b_k$.
The parameters of an HNE composed of $N=2^B$ models are collectively  denoted as $\Theta=\{\theta^0_{1},\dots, \theta^b_{1:2^b}, \dots, \theta^B_{1:2^B}\}$.

\mypar{Anytime inference} 
The success of ensemble learning is due to the reduction in variance by averaging the predictions of different models. 
The expected 
resulting improvements in accuracy are therefore monotonic in  the number of  models in the ensemble. Given that the models in the ensemble can be evaluated  sequentially, the speed-accuracy trade-off 
can be controlled  by choosing how many models to evaluate to approximate the full ensemble output. 
In the case of HNE, this is achieved by evaluating only a
subset of the paths from the root to the leafs, see \fig{intro_figure}. 
More formally, we can choose any value $b \in \{0,1,\dots,B\}$ and compute the ensemble output using a subset of $N^\prime=2^b$ networks as
\begin{equation}
\mathbf{y}^b = \frac{1}{2^b} \sum_{n=1}^{2^b} F (\mathbf{x}; {\theta_n}).
\label{eq:adapt_inference}
\end{equation}
The evaluated subset of $N'=2^b$ networks  is obtained by traversing the binary tree structure in a depth-first manner, where the first evaluated leaf model is always the same.
Thus, we evaluate the first branch, as well as all the other \mbox{$2^b-1$} branches that share the last $b$ blocks with this branch. See \fig{intro_figure}.

\subsection{Computational Complexity}
\label{sec:comp_complexity}
\mypar{Hierarchical vs.\ independent networks}
We analyse the inference complexity of a HNE compared to an ensemble composed of  independent networks. By assuming that functions $f^b(\cdot)$ require the same number of operations, $C$, for all  $b$, the complexity of evaluating all the networks in a HNE is $T_\textrm{HNE} = (2^{B+1}-1)C$, where $B+1$ is the total number of blocks in each model, from root to leaf. This quantity is proportional to $2^{B+1}-1$, which is the total number of nodes in a binary-tree of depth $B$. On the other hand,  an ensemble composed by $N$ networks with independent parameters has an inference cost of $T_\textrm{Ind} = (B+1)NC$.

\begin{figure}
  \begin{center}
    \includegraphics[width=0.75\linewidth]{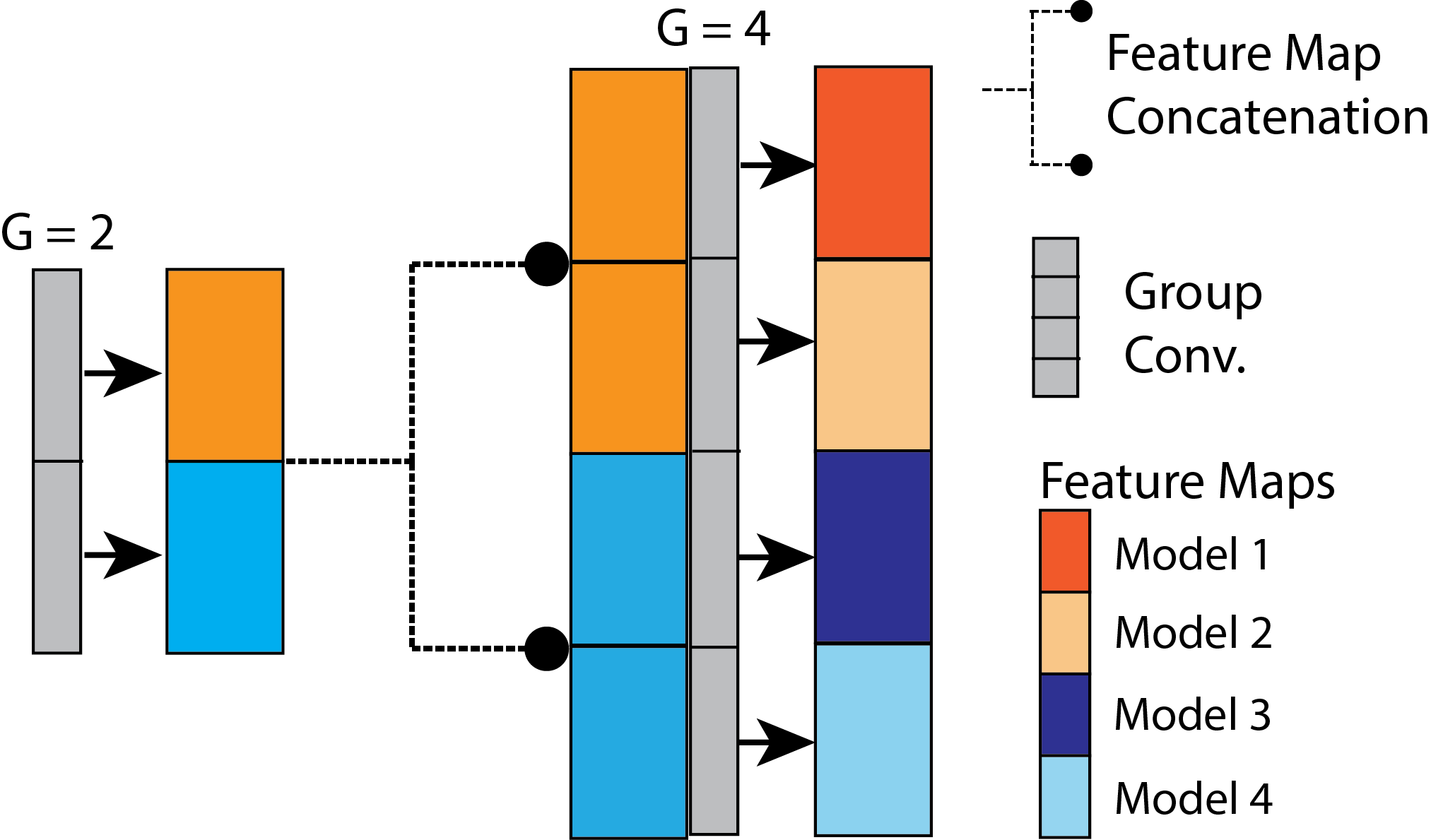}
  \end{center}
  \caption{Efficient HNE implementation using group convolutions. 
Feature maps generated by different branches in the tree are stored in contiguous memory. 
Using group convolutions, the branch outputs can be computed in parallel. When a branch is split, the feature maps are replicated along the channel dimension and the number of groups for the next convolution is doubled.
}
\label{fig:group_conv}
\end{figure}

Considering the same number of models in the ensemble for both approaches,  $N=2^B$,  the ratio between the previous time complexities  is defined by:
\begin{equation}
    \label{eq:coml_relation}
    R = \frac{T_\textrm{Ind}}{T_\textrm{HNE}} = \frac{B+1}{2 -2^{-B}}.
\end{equation}

For $B\!=\!0$,  both independent  and hierarchical ensembles reduce to a single model, and have the same computational complexity ($R\!=\!1$). 
When the number of  models is increased ($B>0$), the second term in the denominator becomes negligible, and the 
speed-up of HNE \textit{w.r.t} to an independent ensemble increases linearly in $B$, with $R\approx  (B+1)/2$. This linear speed-up is important since this is what makes larger ensembles, which enjoy improved accuracy, computationally more affordable using our approach as compared to using  independent ensembles.

\mypar{Efficient HNE implementation}  
Despite the theoretical reduction of the inference complexity, a naive implementation where the individual network outputs are computed sequentially does not allow to fully exploit the  parallelization  provided by  GPUs.
Fortunately, the evaluation of the different networks in the HNE can  be parallelized by means of group convolutions~\cite{howard2017mobilenets,xie2017aggregated}, 
where different sets of input channels are used to compute an independent set of outputs, see \fig{group_conv}. Compared to sequential model evaluation, this strategy allows to drastically reduce training time given that all the models can be computed with a single forward-pass.

\subsection{HNE Optimization}
\label{sec:hne_losses}

Given a training set $D=\{(\mathbf{x}_,\mathbf{y})_{1:M}\}$ composed of $M$ sample and label pairs, HNE parameters are optimized by minimizing a loss function for each individual network as
\begin{equation}
\label{eq:loss_function}
\mathcal{L}^\textrm{I}(\Theta) = \sum_{n=1}^N \sum_{m=1}^M \ell(F(\mathbf{x}_m; {\theta_n}) ,\mathbf{y}_m),
\end{equation}
where $\ell(\cdot,\cdot)$ is the cross-entropy loss comparing ground-truth labels with network outputs. 

A drawback of the loss in \Eq{loss_function} is that it is symmetric among the different models in the ensemble. 
Notably, it ignores the hierarchical structure of the  sub-trees that are used to compose smaller sub-ensembles for adaptive inference complexity. 
To address this limitation, we can optimize a loss that measures the accuracy  of the different sub-trees corresponding to the evaluation of an increasing number of networks in the ensemble: 
\begin{equation}
\mathcal{L}^\textrm{S}(\Theta) = \sum_{b=0}^{B} \sum_{m=1}^M \ell(\mathbf{y}^b_m ,\mathbf{y}_m),
\label{eq:h_loss}
\end{equation}
where $\mathbf{y}_m^b$ is defined in Eq. (\ref{eq:adapt_inference}).
Despite the apparent advantages of replacing \Eq{loss_function} by \Eq{h_loss} during learning, 
we empirically show that this strategy generally produces worse results. 
The reason is that \Eq{h_loss} impedes the branches to behave as an ensemble of independent networks. 
Instead, the different models  tend to co-adapt in order to minimize the training error. As a consequence, averaging their outputs does not reduce the variance over test data predictions. 
To effectively exploit the hierarchical structure of HNE outputs during learning, we propose an alternative approach below.

\subsection{Hierarchical Distillation}

Previous work  on network ensembles have explored the use of co-distillation \cite{anil2018large,song2018collaborative}. These methods attempt to transfer the ensemble knowledge to the individual models by introducing an auxiliary distillation loss for each network:
\begin{equation}
\label{eq:dist_loss}
\mathcal{L}^\textrm{D}(\Theta) = \sum_{n=1}^N \sum_{m=1}^M \ell(F_n(\mathbf{x}_m; {\theta_n}) ,\mathbf{y}^E_m),
\end{equation}
where 
$\mathbf{y}^E = \frac{1}{N} \sum_{n=1}^{N} F_n (\mathbf{x}_m; {\theta_n})$
is the ensemble output for sample $\mathbf{x}_m$. 
The cross-entropy loss $\ell(\cdot,\cdot)$ compares the network outputs with the soft-labels generated by using a soft-max function over $\mathbf{y}^E_m$.
During training, the distillation loss is combined with the  cross-entropy loss of \Eq{loss_function} as $(1-\alpha) \mathcal{L}^\textrm{I} + \alpha \mathcal{L}^\textrm{D}$, where $\alpha$ is an hyper-parameter controlling the trade-off between both terms. 
The gradients of $\mathbf{y}^E_m$ \textit{w.r.t} $\Theta$ parameters are not back-propagated during optimization.

Whereas this distillation approach boosts the performance of individual models, it has a critical drawback in the context of ensemble learning for anytime inference. 
In particular,  co-distillation encourage all the predictions to be similar to their average. 
As a consequence, the variance between model predictions decreases, limiting the improvement given by combining multiple models in an ensemble. To address this limitation, we propose a novel distillation scheme which we refer to it as ``hierarchical distillation''. 
The core idea is to transfer the knowledge from the full ensemble to the smaller sub-ensembles used for anytime inference in HNE. 
In particular, we minimize:
\begin{equation}
\label{eq:hdist_loss}
\mathcal{L}^\textrm{HD}(\Theta) = \sum_{b=0}^{B-1} \sum_{m=1}^M \ell(\mathbf{y}_m^b ,\mathbf{y}^E_m).
\end{equation}

Different from $\mathcal{L}^\textrm{S}(\cdot)$, our hierarchical distillation loss distills the predictions for each sub-tree towards the full ensemble outputs. 
Additionally,  $\mathcal{L}^\textrm{HD}(\cdot)$ does not force all the independent outputs to be similar to the full ensemble prediction as in standard distillation. 
In contrast, our hierarchical distillation loss encourages the ensemble prediction obtained from a subset of models to match the full ensemble prediction. Therefore, the outputs between the individual models in this subset can be diverse and still minimize the distillation loss, preserving  model diversity and retaining the advantages of averaging multiple networks. Given that the first evaluated model in the tree is fixed, the different sub-ensembles are always composed of the same subset of networks.
As empirically shown in our experiments, the proposed distillation loss slightly reduces the model diversity, 
as compared to training without distillation.
However, the accuracy of individual models tends to be much higher and thus, the ensembles performance is significantly improved.

\section{Experiments}
\label{sec:impl_details}

\begin{figure*}[ht]
\centering
\includegraphics[width=0.9\linewidth]{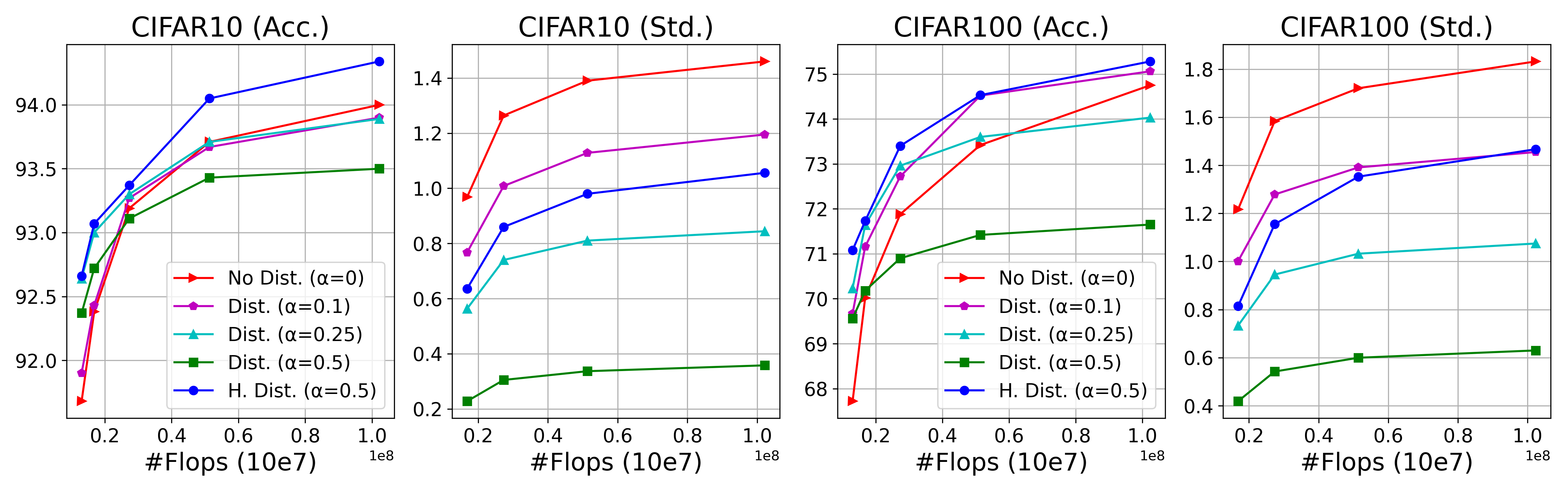}
\caption{Accuracy and standard deviation in logits \textit{vs} FLOPs  for HNE$_{small}$ trained (i) without  distillation, (ii) with  distillation,  and (iii) with our hierarchical distillation. Curves represent results for ensembles of size 1 up to 16.}
\label{fig:distillation_experiment}
\end{figure*}

\begin{figure*}[ht]
\centering
\includegraphics[width=0.9\linewidth]{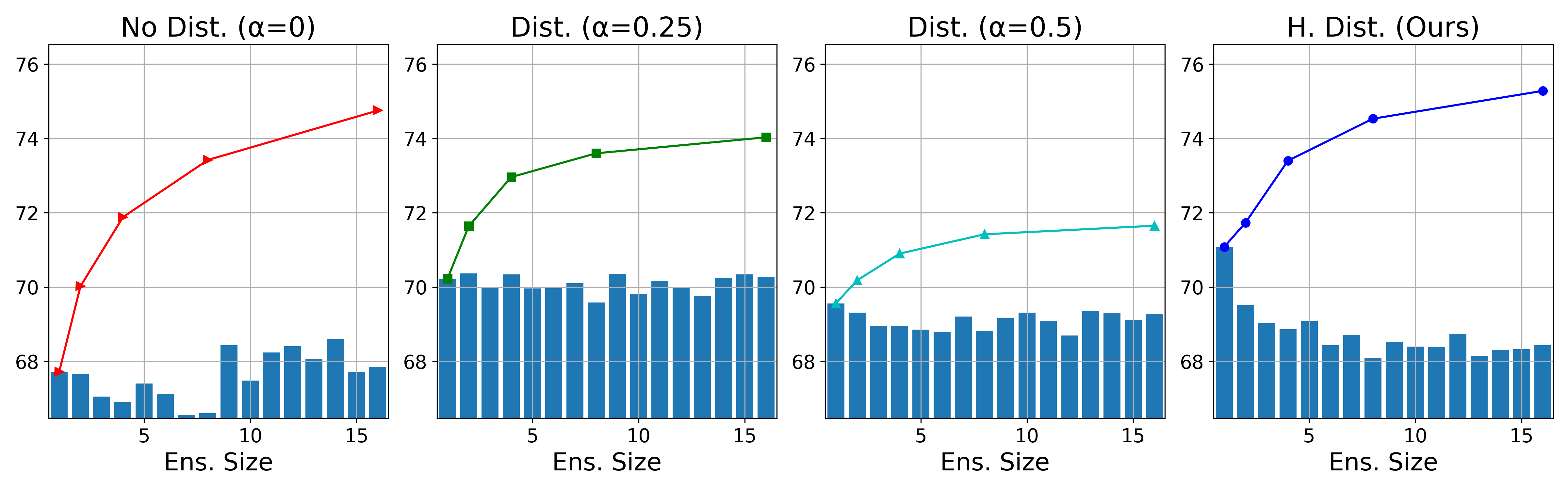}
\caption{
Results on CIFAR-100 for HNE$_{small}$ trained without distillation, standard distillation and the proposed hierarchical distillation. Curves indicate the performance of  ensembles of different sizes. Bars depict the accuracy of  individual models.
}
\label{fig:results_distillation_models}
\end{figure*}

\mypar{Datasets} We  use the  CIFAR-10/100 ~\cite{krizhevsky2009learning} and  ImageNet~\cite{ILSVRC15} datasets. CIFAR-10/100 contain 50k train and 10k test images from 10 and 100 classes, respectively. Following standard protocols \cite{he2016deep}, we pre-process the images by normalizing their mean and standard-deviation for each color channel. Additionally, during training we use a data augmentation process where we extract random crops of 32$\times$32 after applying a $4$-pixel zero padding to the original image or its horizontal flip. 
Imagenet is composed by 1.2M and 50k high-resolution images for training and validation, respectively, labelled across 1,000 different categories. 
We use the standard protocol during evaluation resizing the image and extracting a center crop of $224\times224$ \cite{he2016deep}. For training, we apply the same data augmentation process as in \cite{yang2020resolution,huang2017multi}.
For both datasets we report the classification accuracy.

\begin{table}[t]
\centering
\resizebox{0.9\linewidth}{!}{%
\begin{tabular}{llcccccc}
\multicolumn{1}{l}{} &  & \multicolumn{1}{l}{} & \multicolumn{5}{c}{\textbf{\# models evaluated for inference}} \\ \cline{4-8} 
\multicolumn{1}{l}{} &  & \multicolumn{1}{l}{} & \textbf{1} & \textbf{2} & \textbf{4} & \textbf{8} & \textbf{16} \\ \hline
\multicolumn{1}{c|}{} & \multicolumn{1}{l|}{} & $\mathcal{L^I}$ & 91.7 & 92.4 & 93.2 & 93.7 & 94.0 \\
\multicolumn{1}{c|}{} & \multicolumn{1}{l|}{} & \cellcolor[HTML]{EFEFEF}$\mathcal{L^S}$ & \cellcolor[HTML]{EFEFEF}91.6 & \cellcolor[HTML]{EFEFEF}92.1 & \cellcolor[HTML]{EFEFEF}92.7 & \cellcolor[HTML]{EFEFEF}92.9 & \cellcolor[HTML]{EFEFEF}93.3 \\
\multicolumn{1}{c|}{} & \multicolumn{1}{l|}{\multirow{-4}{*}{\textbf{C10}}} & \cellcolor[HTML]{EFEFEF}$\mathcal{L}^{HD}$ & \cellcolor[HTML]{EFEFEF}\textbf{92.7} & \cellcolor[HTML]{EFEFEF}\textbf{93.1} & \cellcolor[HTML]{EFEFEF}\textbf{93.4} & \cellcolor[HTML]{EFEFEF}\textbf{94.1} & \cellcolor[HTML]{EFEFEF}\textbf{94.3} \\ \cline{2-8} 
\multicolumn{1}{c|}{} & \multicolumn{1}{l|}{} & $\mathcal{L^I}$ & 67.7 & 70.0 & 71.9 & 73.4 & 74.8 \\
\multicolumn{1}{c|}{} & \multicolumn{1}{l|}{} & \cellcolor[HTML]{EFEFEF}{\color[HTML]{000000} $\mathcal{L^S}$} & \cellcolor[HTML]{EFEFEF}{\color[HTML]{000000} 65.0} & \cellcolor[HTML]{EFEFEF}{\color[HTML]{000000} 66.3} & \cellcolor[HTML]{EFEFEF}{\color[HTML]{000000} 68.3} & \cellcolor[HTML]{EFEFEF}{\color[HTML]{000000} 69.6} & \cellcolor[HTML]{EFEFEF}{\color[HTML]{000000} 72.6} \\
\multicolumn{1}{c|}{\multirow{-8}{*}{\begin{tabular}[c]{@{}c@{}}$\mathbf{HNE}_{small}$\\ (N=16)\end{tabular}}} & \multicolumn{1}{l|}{\multirow{-4}{*}{\textbf{C100}}} & \cellcolor[HTML]{EFEFEF}$\mathcal{L}^{HD}$ & \cellcolor[HTML]{EFEFEF}\textbf{71.1} & \cellcolor[HTML]{EFEFEF}\textbf{71.7} & \cellcolor[HTML]{EFEFEF}\textbf{73.4} & \cellcolor[HTML]{EFEFEF}\textbf{74.5} & \cellcolor[HTML]{EFEFEF}\textbf{75.3} \\ \hline
\multicolumn{1}{c|}{} & \multicolumn{1}{l|}{} & $\mathcal{L^I}$ & 93.6 & 94.2 & 94.8 & 95.0 & 95.2 \\
\multicolumn{1}{c|}{} & \multicolumn{1}{l|}{} & \cellcolor[HTML]{EFEFEF}$\mathcal{L^S}$ & \cellcolor[HTML]{EFEFEF}92.9 & \cellcolor[HTML]{EFEFEF}93.6 & \cellcolor[HTML]{EFEFEF}93.5 & \cellcolor[HTML]{EFEFEF}94.1 & \cellcolor[HTML]{EFEFEF}94.4 \\
\multicolumn{1}{c|}{} & \multicolumn{1}{l|}{\multirow{-4}{*}{\textbf{C10}}} & \cellcolor[HTML]{EFEFEF}{\color[HTML]{000000} $\mathcal{L}^{HD}$} & \cellcolor[HTML]{EFEFEF}{\color[HTML]{000000} \textbf{94.6}} & \cellcolor[HTML]{EFEFEF}{\color[HTML]{000000} \textbf{94.9}} & \cellcolor[HTML]{EFEFEF}{\color[HTML]{000000} \textbf{95.1}} & \cellcolor[HTML]{EFEFEF}{\color[HTML]{000000} \textbf{95.5}} & \cellcolor[HTML]{EFEFEF}{\color[HTML]{000000} \textbf{95.6}} \\ \cline{2-8} 
\multicolumn{1}{c|}{} & \multicolumn{1}{l|}{} & $\mathcal{L^I}$ & 73.5 & 75.4 & 77.5 & \textbf{79.0} & 79.7 \\
\multicolumn{1}{c|}{} & \multicolumn{1}{l|}{} & \cellcolor[HTML]{EFEFEF}$\mathcal{L^S}$ & \cellcolor[HTML]{EFEFEF}70.7 & \cellcolor[HTML]{EFEFEF}73.3 & \cellcolor[HTML]{EFEFEF}74.0 & \cellcolor[HTML]{EFEFEF}75.7 & \cellcolor[HTML]{EFEFEF}76.8 \\
\multicolumn{1}{c|}{\multirow{-8}{*}{\begin{tabular}[c]{@{}c@{}}$\mathbf{HNE}$\\ (N=16)\end{tabular}}} & \multicolumn{1}{l|}{\multirow{-4}{*}{\textbf{C100}}} & \cellcolor[HTML]{EFEFEF}$\mathcal{L}^{HD}$ & \cellcolor[HTML]{EFEFEF}\textbf{76.1} & \cellcolor[HTML]{EFEFEF}\textbf{77.2} & \cellcolor[HTML]{EFEFEF}\textbf{78.0} & \cellcolor[HTML]{EFEFEF}\textbf{79.0} & \cellcolor[HTML]{EFEFEF}\textbf{79.8} \\ \hline
\end{tabular}%
}
\caption{Accuracy  of HNE and HNE$_{small}$ embedding 16 different networks on CIFAR. Columns correspond to the number of models evaluated during inference.}
\label{tab:loss_evaluation}

\end{table}

\mypar{Base architectures}
We implement HNE with   commonly used   architectures.
For CIFAR-10/100, we use a variant of ResNet \cite{he2016deep}, composed of a sequence of residual convolutional layers with bottlenecks. 
We employ depth-wise instead of regular convolutions to reduce computational complexity. 
We generate a HNE with a total five blocks embedding an ensemble of $N=16$ CNNs. 
We  report results for the base ResNet  architecture, as well as a version  where for all layers we divided the number of   feature channels by two ($\text{HNE}_{small}$). 
This  provides a more complete evaluation by adding a regime where inference is extremely efficient.

\begin{figure*}
\begin{center}
\includegraphics[trim=140 0 140 20, clip, width=0.9\linewidth]{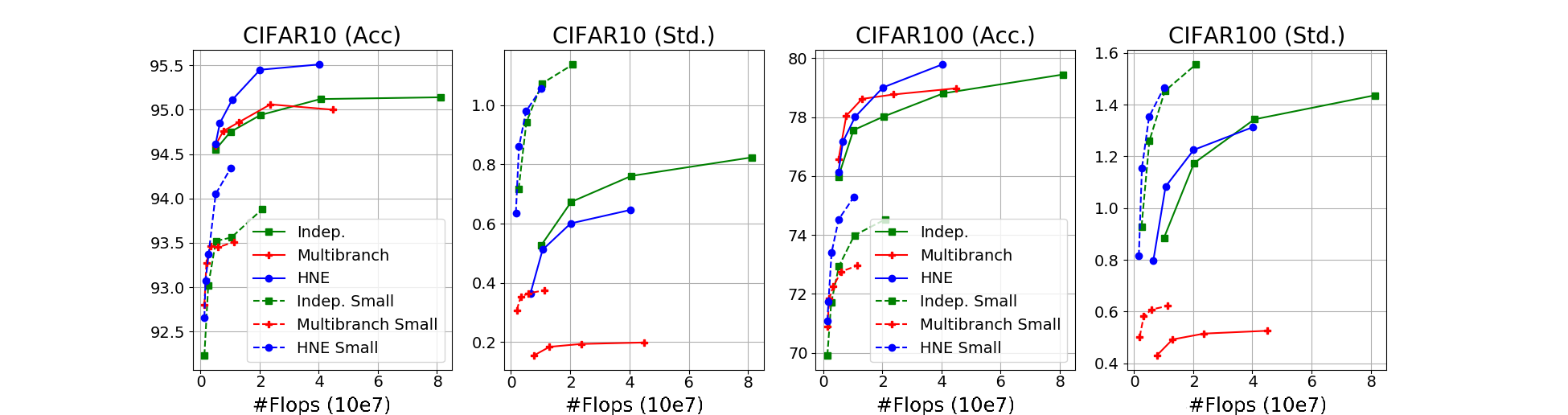}
\end{center}
\caption{Results on CIFAR-100 for HNE$_{small}$ using  different ensemble architectures: (i) fully-independent networks, (ii) multi-branch architecture with shared backbone, (iii) our proposed HNEs.}
\label{fig:structure_experiment}
\end{figure*}

For ImageNet, we implement an HNE based on MobileNet v2 \cite{sandler2018mobilenetv2}, which uses inverted residual layers and depth-wise convolutions as main building blocks. In this case, we also use five blocks generating $N=16$ different networks. 
In the supplementary material  we present a detailed description of our HNE implementation using ResNet and MobileNetv2 and  provide all the training hyper-parameters. The particular design choices for both architectures are set to produce a similar computational complexity as previous methods to which we compare. We provide a Pytorch implementation of HNE in order to reproduce the main results in our paper \footnote{\url{https://gitlab.com/adriaruizo/dhne-aaai21}}.

\mypar{Inference complexity.} Following~\cite{huang2017multi,ruiz2019adaptative,zhang2018graph}, we evaluate the computational complexity of inference in the models according to the number of floating-point operations (FLOPs). The advantages of this metric are that it is independent from differences in hardware and implementations, and strongly correlated with 
wall-clock  inference time.

\subsection{Ablation Study on CIFAR-10/100}
\label{sec:ablation}

\mypar{Optimizing HNE}
In order to understand the advantages of our hierarchical distillation approach, 
we compare three different alternative objectives to train HNE:
(i) the independent loss across models, $\mathcal{L}^\textrm{I}$ in \Eq{loss_function},
(ii) the structured loss maximizing the accuracy of nested ensembles, $\mathcal{L}^\textrm{S}$ in \Eq{h_loss},
and 
(iii)  our hierarchical distillation loss, $\mathcal{L}^\textrm{HD}$ in \Eq{hdist_loss}, which is combined with $\mathcal{L}_\textrm{I}$.

As shown in \tab{loss_evaluation}, $\mathcal{L}^\textrm{I}$ provides better performance as compared to training with  $\mathcal{L}^\textrm{S}$. 
The reason is that $\mathcal{L}^\textrm{S}$ encourages  individual model outputs to co-adapt in order to minimize the training error. However, as the different networks are not trained independently, the variance reduction resulting from averaging multiple models in an ensemble is lost, causing a performance drop on test data. 
Using our hierarchical distillation loss, however, consistently outperforms the alternatives in all the evaluated ensemble sizes, both architectures, and on both datasets. This is because our approach preserves the advantages of averaging multiple independent models, at the same time that the performance of hierarchical ensembles is increased via distillation.

\mypar{Comparing distillation approaches} 
After demonstrating the effectiveness of our distillation method to boost the performance of hierarchical ensembles, we evaluate its advantages \textit{w.r.t}  standard distillation. 
For this purpose, we train HNE using the loss $\mathcal{L}^\textrm{D}$ of \Eq{dist_loss}, as~\cite{song2018collaborative}. 
For standard distillation we evaluate a range of values for $\alpha$ to mix the distillation and cross-entropy loss in order to analyze the  impact on accuracy and model diversity. 
The latter is measured as the std.\ deviation in the logits of the evaluated models, averaged across all classes and test samples.
In \fig{distillation_experiment} we report both the accuracy and logit standard deviation on the test set. For limited space reasons, we only show results for HNE$_{small}$. The corresponding figure for the bigger HNE model can be found in supplementary material.

Consider standard distillation with a high weight on the distillation loss  ($\alpha=0.5$). As expected, the performance of small ensembles  is improved \textit{w.r.t}  training without  distillation. 
For larger ensembles, however,  the accuracy tends to be  significantly lower  compared to not using distillation. 
This is due to the reduction in diversity among the models induced by the standard distillation loss.
This effect can be controlled by reducing the weight $\alpha$, but smaller values (less distillation) compromise the accuracy of small ensembles. 
In contrast, our hierarchical distillation achieves the best FLOPs-accuracy trade-offs for all ensemble sizes and datasets. For small ensemble sizes, our hierarchical distillation obtains similar or better accuracy than standard distillation. For large ensemble sizes, our  approach  significantly improves over standard distillation, and the accuracy is comparable or better than those obtained without distillation. These results clearly shows the advantages of hierarchical distillation for any-time inference. The reason is that, in this setting, the goal is not only in to optimize the accuracy for a given FLOP count, but to jointly boost the performance for all the possible ensemble sizes.

\begin{figure*}
\begin{center}
\begin{minipage}{.28\textwidth}
  \centering
  \includegraphics[width=\linewidth]{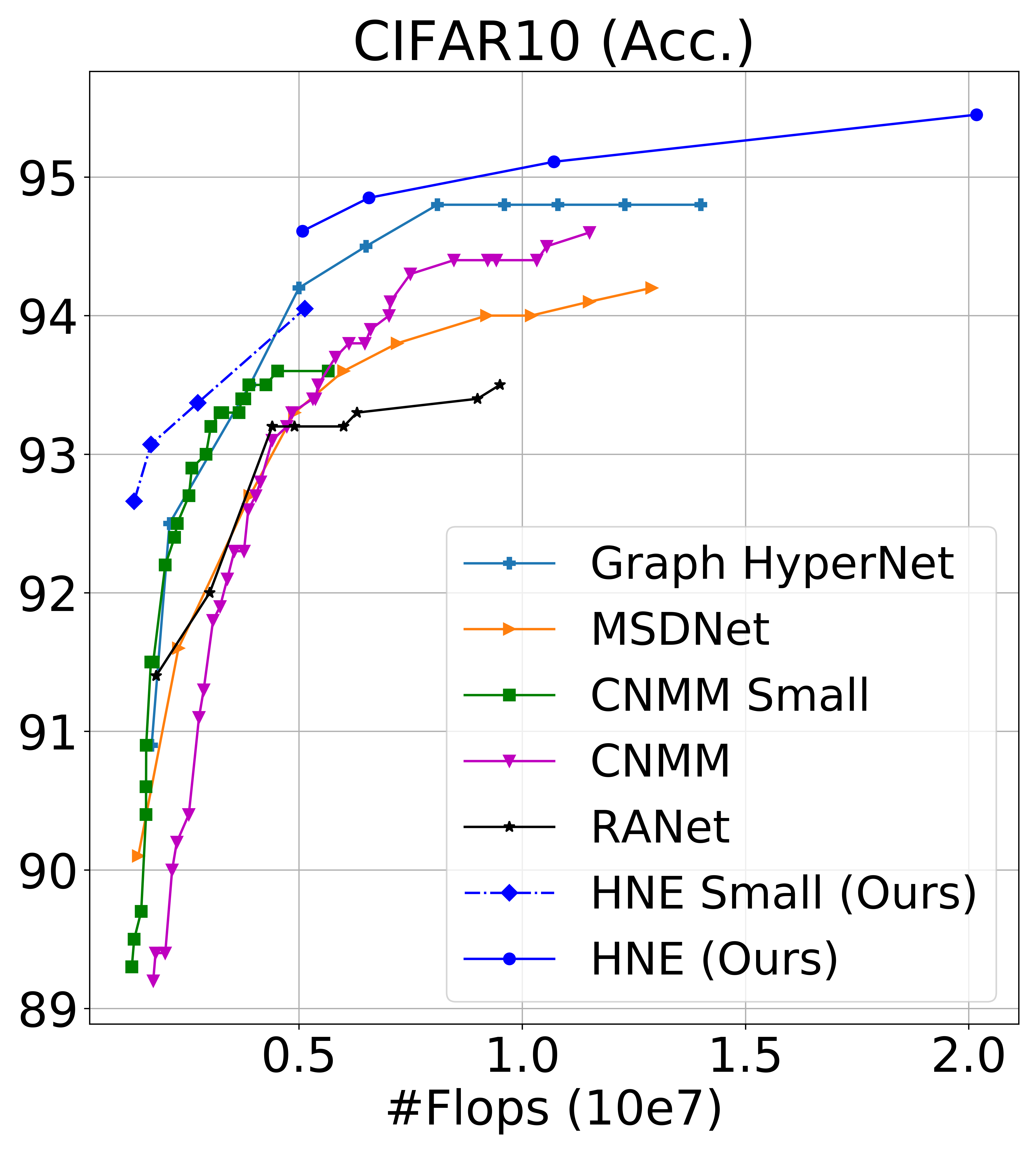}

\end{minipage}%
\begin{minipage}{.28\textwidth}
  \centering
  \includegraphics[width=\linewidth]{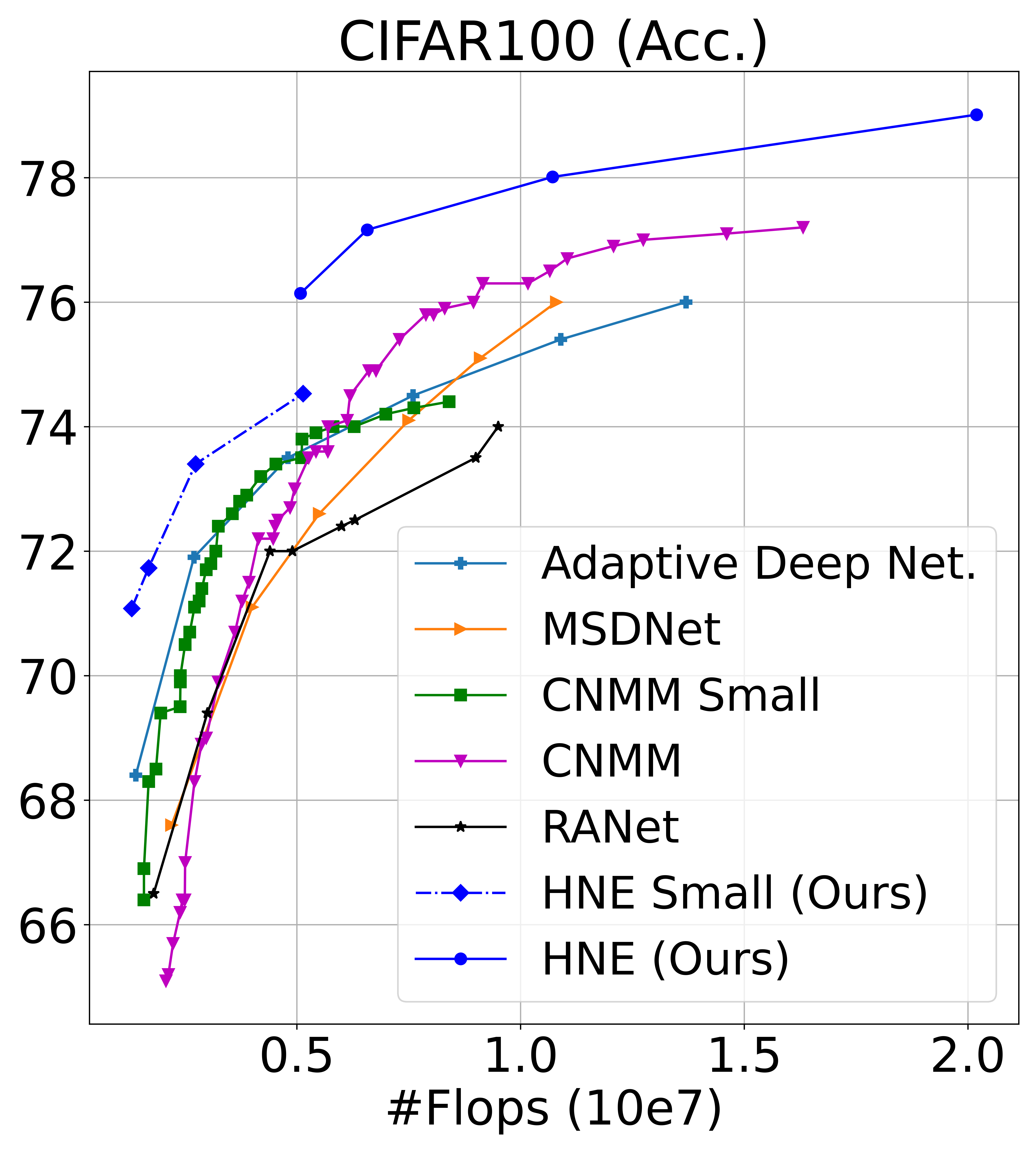}

\end{minipage}
\begin{minipage}{.28\textwidth}
  \centering
  \includegraphics[width=\linewidth]{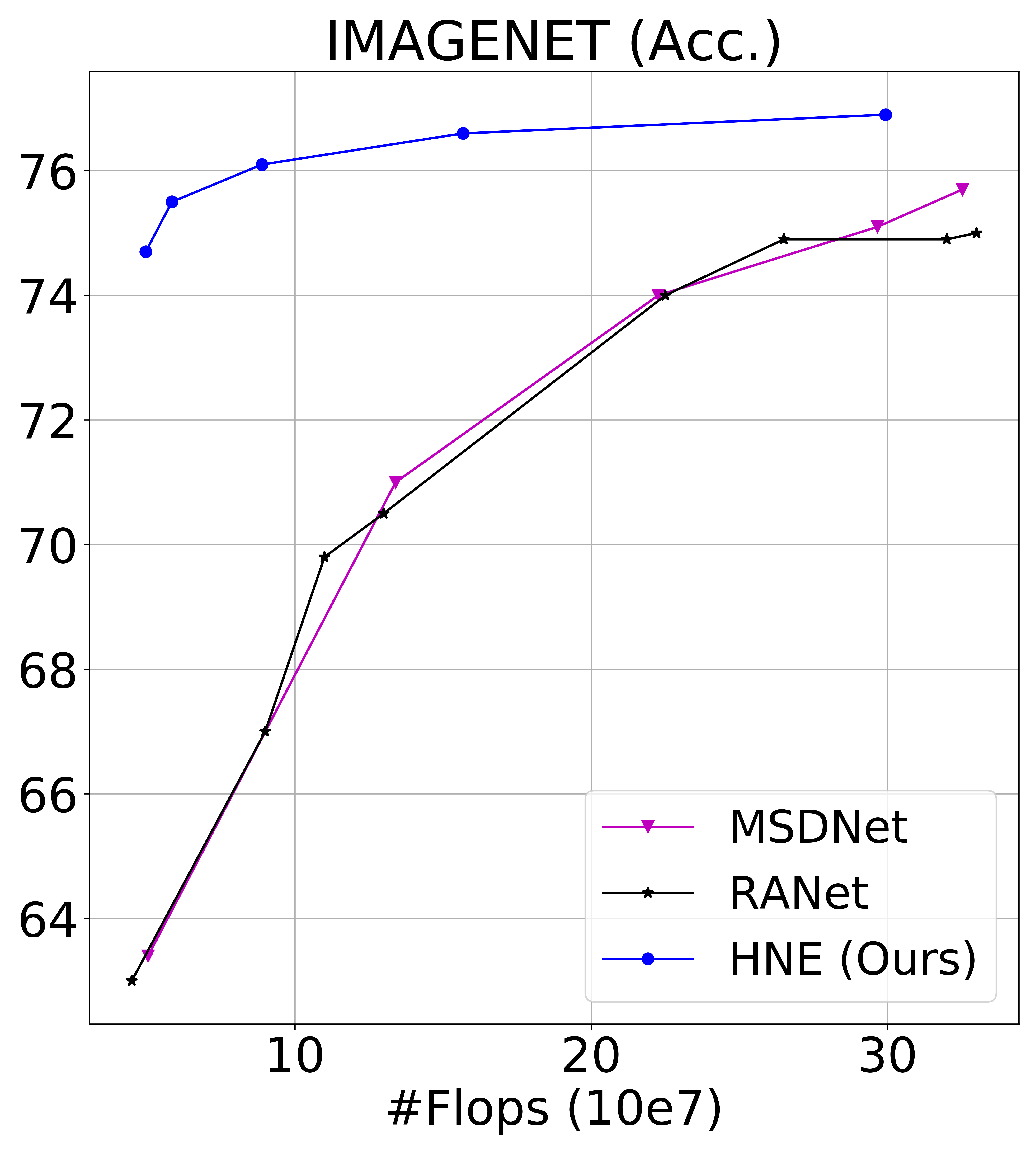}

\end{minipage}
\end{center}
\caption{Comparison of HNE with the state of the art. 
Each curve corresponds to a single model with  anytime inference.
}
\label{fig:soa_results}
\end{figure*}

\mypar{Analysis of individual network accuracies} To provide additional insight in the previous results, \fig{results_distillation_models} depicts the performance of HNE for  different ensemble sizes used during inference (curves), and the accuracy of the individual networks in the ensemble (bars).
Comparing the results without distillation (first col.) to standard distillation (second and third cols.), we   make two observations. First, standard distillation significantly increases the accuracy of the individual models. This is expected because the knowledge from the complete ensemble is transferred to each network independently. 
Second,  when using standard distillation, performance tends to be lower than HNE trained without distillation when the number of models in the ensemble is increased. 
Both phenomena are  explained by the tendency of  standard distillation to decrease the diversity between the individual models. As a consequence, the gains obtained by combining a large number of networks are reduced even though they are individually more accurate. 

The  results for our Hierarchical Distillation (last col.) clearly show its advantages with respect to the alternative approaches. We can observe that the accuracy of the first model is better than in HNE trained without distillation, and also significantly higher than the accuracy of the model obtained by standard distillation. 
The reason is that the ensemble knowledge is directly transferred to the predictions of the first network in the hierarchical structure. 
The performance of the other individual networks in the ensemble tends to be lower than when training  with standard distillation. 
The improvement obtained by ensembling their predictions is, however, significantly larger. 
This is because hierarchical distillation preserves the diversity between networks, compensating the lower accuracy of the individual models.

\mypar{Hierarchical parameter sharing} 
We compare the performance obtained by HNE and an ensemble of independent networks that are not organized in a tree structure, using the same base architecture. 
We also compare to multi-branch architectures~\cite{lan2018knowledge,lee2015m}. 
In the latter case, the complexity is reduced by sharing the same ``backbone'' for the first blocks for all models, and then splits in one step to $N$ independent branches for the subsequent blocks. 
For HNE we use five different blocks with $N=16$.
To achieve architectures in a similar FLOP range, we use the first three blocks as a backbone and implement 16 independent branches for the last two blocks. 
We use our hierarchical distillation loss in all cases. 

In \fig{structure_experiment} we  report both the accuracy (Acc) and the standard deviation of logits (Std).
We  observe that HNE obtains significantly better results than the independent networks. 
Moreover, the hierarchical structure allows to significantly reduce the computational cost for large ensemble sizes. 
For small ensembles on CIFAR-100 the multi-branch models obtain slightly better accuracy than HNE. 
In all other settings,  HNE achieve similar or significantly better accuracy, especially for larger ensembles. The results can again be understood by observing the diversity across models.
HNE and independent models have similar diversity, while in the case of the multi-branch ensemble diversity is significantly lower. This shows the importance of using different parameters in early-blocks.

\subsection{Comparison with the  state of the art}
\label{sec:soa}

\mypar{CIFAR-10/100} 
We compare the performance of HNE trained with hierarchical distillation with state-of-the-art approaches for anytime inference: Multi-scale DenseNets \cite{huang2017multi}, Resolution Adaptive Networks \cite{yang2020resolution}, Graph Hyper-Networks \cite{zhang2018graph}, Deep Adaptive Networks \cite{li2019improved}, and Convolutional Neural Mixture Models \cite{ruiz2019adaptative}. We  report HNE results  for ensembles of sizes up to eight, in order to provide a maximum FLOP count  similar to the compared methods ($<$250M). Results in \fig{soa_results} show that for both datasets HNE significantly outperforms previous approaches across all the FLOP range. 

\mypar{ImageNet} We compare our method with Multi-Scale Densenets \cite{huang2017multi}, and Resolution Adaptive Networks \cite{yang2020resolution}. To the best of our knowledge these works have reported state-of-the-art performance on ImageNet for anytime inference. 
The results in \fig{soa_results} show that our  HNE achieves better  accuracy than the compared methods across all inference complexities. Compared to the best baseline, our method achieves an accuracy improvement across the different FLOP ranges between 1.5\% and 11\%. This is a significant performance boost given the difficulty and large-scale nature of ImageNet. Additionally, note that the minimum FLOP count for HNE and the compared models are similar. Whereas HNE needs a full pass over a single base model to provide an initial output, the compared approaches based on intermediate classifiers also require to compute all the intermediate network activations up to the first classifier.

\section{Conclusions}
In this paper we proposed Hierarchical Neural Ensembles (HNE), a framework to design deep models with anytime inference. 
In addition, we  introduced a novel hierarchical distillation approach adapted to the structure of HNE. 
Compared to previous deep models with anytime inference, we have reported state-of-the-art compute-accuracy trade-offs on CIFAR-10/100 and ImageNet.
While we have demonstrated the effectiveness of our framework in the context of CNNs for image classification, our approach is generic and can be used to build ensembles of other types of deep networks for different tasks and domains. In particular, HNE can be applied to any base model and network branching. This property allows to design anytime models adapted to different computational constraints such as the maximum and minimum FLOP count or the number of desired operating points. This flexibility allows our framework to be combined with other approaches for efficient inference such as network compression or neural architecture search.

\clearpage
\section*{Potential Ethical Impact}
Our work contributes to an ongoing broad effort to develop energy-efficient machine learning approaches. Recent works have analysed the environmental impact of the massive use of deep networks in research and industry  \cite{strubell2019energy,henderson2020towards}. These works have pointed out the large "carbon footprint" of these type of models and the need to start focusing on their efficiency apart from their raw accuracy \cite{schwartz2019green}. Our work contributes to the development of efficient models with low energy consumption, which helps to deploy large-scale artificial intelligence systems in a environmentally friendly manner.

\section*
{Acknowledgments}
This work has been partially supported by the project IPLAM PCI2019-103386. Adria Ruiz acknowledges financial support from MICINN (Spain) through the program Juan de la Cierva.
\bibliography{paper}

\begin{thebibliography}{61}
\providecommand{\natexlab}[1]{#1}
\providecommand{\url}[1]{\texttt{#1}}
\providecommand{\urlprefix}{URL }
\expandafter\ifx\csname urlstyle\endcsname\relax
  \providecommand{\doi}[1]{doi:\discretionary{}{}{}#1}\else
  \providecommand{\doi}{doi:\discretionary{}{}{}\begingroup
  \urlstyle{rm}\Url}\fi

\bibitem[{Anil et~al.(2018)Anil, Pereyra, Passos, Ormandi, Dahl, and
  Hinton}]{anil2018large}
Anil, R.; Pereyra, G.; Passos, A.; Ormandi, R.; Dahl, G.~E.; and Hinton, G.~E.
  2018.
\newblock Large scale distributed neural network training through online
  distillation.
\newblock \emph{ICLR} .

\bibitem[{Ba and Caruana(2014)}]{ba2014deep}
Ba, J.; and Caruana, R. 2014.
\newblock Do deep nets really need to be deep?
\newblock In \emph{NIPS}.

\bibitem[{Bhardwaj et~al.(2019)Bhardwaj, Lin, Sartor, and
  Marculescu}]{bhardwaj2019memory}
Bhardwaj, K.; Lin, C.-Y.; Sartor, A.; and Marculescu, R. 2019.
\newblock Memory-and communication-aware model compression for distributed deep
  learning inference on iot.
\newblock \emph{ACM Transactions on Embedded Computing Systems (TECS)} 18(5s):
  1--22.

\bibitem[{Bolukbasi et~al.(2017)Bolukbasi, Wang, Dekel, and
  Saligrama}]{bolukbasi2017adaptive}
Bolukbasi, T.; Wang, J.; Dekel, O.; and Saligrama, V. 2017.
\newblock Adaptive neural networks for efficient inference.
\newblock In \emph{ICML}.

\bibitem[{Breiman(1996)}]{breiman1996bagging}
Breiman, L. 1996.
\newblock Bagging predictors.
\newblock \emph{Machine learning} .

\bibitem[{Cai et~al.(2020)Cai, Gan, Wang, Zhang, and Han}]{cai2020once}
Cai, H.; Gan, C.; Wang, T.; Zhang, Z.; and Han, S. 2020.
\newblock Once for All: Train One Network and Specialize it for Efficient
  Deployment.
\newblock In \emph{ICLR}.

\bibitem[{Chen and Yao(2009)}]{chen2009regularized}
Chen, H.; and Yao, X. 2009.
\newblock Regularized negative correlation learning for neural network
  ensembles.
\newblock \emph{IEEE Transactions on Neural Networks} .

\bibitem[{Elbayad et~al.(2020)Elbayad, Gu, Grave, and Auli}]{elbayad20iclr}
Elbayad, M.; Gu, J.; Grave, E.; and Auli, M. 2020.
\newblock Depth-Adaptive Transformer.
\newblock In \emph{ICLR}.

\bibitem[{Furlanello et~al.(2018)Furlanello, Lipton, Tschannen, Itti, and
  Anandkumar}]{furlanello2018born}
Furlanello, T.; Lipton, Z.~C.; Tschannen, M.; Itti, L.; and Anandkumar, A.
  2018.
\newblock Born again neural networks.
\newblock In \emph{ICML}.

\bibitem[{Geiger et~al.(2020)Geiger, Jacot, Spigler, Gabriel, Sagun, d'Ascoli,
  Biroli, Hongler, and Wyart}]{geiger20jsm}
Geiger, M.; Jacot, A.; Spigler, S.; Gabriel, F.; Sagun, L.; d'Ascoli, S.;
  Biroli, G.; Hongler, C.; and Wyart, M. 2020.
\newblock Scaling description of generalization with number of parameters in
  deep learning.
\newblock \emph{Journal of Statistical Mechanics: Theory and Experiment}
  2020(2).

\bibitem[{Hansen and Salamon(1990)}]{hansen90pami}
Hansen, L.; and Salamon, P. 1990.
\newblock Neural Network Ensembles.
\newblock \emph{PAMI} 12(10).

\bibitem[{He et~al.(2016)He, Zhang, Ren, and Sun}]{he2016deep}
He, K.; Zhang, X.; Ren, S.; and Sun, J. 2016.
\newblock Deep residual learning for image recognition.
\newblock In \emph{CVPR}.

\bibitem[{Henderson et~al.(2020)Henderson, Hu, Romoff, Brunskill, Jurafsky, and
  Pineau}]{henderson2020towards}
Henderson, P.; Hu, J.; Romoff, J.; Brunskill, E.; Jurafsky, D.; and Pineau, J.
  2020.
\newblock Towards the Systematic Reporting of the Energy and Carbon Footprints
  of Machine Learning.
\newblock \emph{arXiv preprint arXiv:2002.05651} .

\bibitem[{Hinton, Vinyals, and Dean(2015)}]{hinton2015distilling}
Hinton, G.; Vinyals, O.; and Dean, J. 2015.
\newblock Distilling the knowledge in a neural network.
\newblock \emph{arXiv preprint arXiv:1503.02531} .

\bibitem[{Howard et~al.(2019)Howard, Sandler, Chu, Chen, Chen, Tan, Wang, Zhu,
  Pang, Vasudevan et~al.}]{howard2019searching}
Howard, A.; Sandler, M.; Chu, G.; Chen, L.-C.; Chen, B.; Tan, M.; Wang, W.;
  Zhu, Y.; Pang, R.; Vasudevan, V.; et~al. 2019.
\newblock Searching for mobilenetv3.
\newblock In \emph{ICCV}.

\bibitem[{Howard et~al.(2017)Howard, Zhu, Chen, Kalenichenko, Wang, Weyand,
  Andreetto, and Adam}]{howard2017mobilenets}
Howard, A.~G.; Zhu, M.; Chen, B.; Kalenichenko, D.; Wang, W.; Weyand, T.;
  Andreetto, M.; and Adam, H. 2017.
\newblock Mobilenets: Efficient convolutional neural networks for mobile vision
  applications.
\newblock \emph{arXiv preprint arXiv:1704.04861} .

\bibitem[{Huang et~al.(2018{\natexlab{a}})Huang, Chen, Li, Wu, van~der Maaten,
  and Weinberger}]{huang2017multi}
Huang, G.; Chen, D.; Li, T.; Wu, F.; van~der Maaten, L.; and Weinberger, K.~Q.
  2018{\natexlab{a}}.
\newblock Multi-scale dense networks for resource efficient image
  classification.
\newblock \emph{ICLR} .

\bibitem[{Huang et~al.(2017)Huang, Li, Pleiss, Liu, Hopcroft, and
  Weinberger}]{huang2017snapshot}
Huang, G.; Li, Y.; Pleiss, G.; Liu, Z.; Hopcroft, J.; and Weinberger, K. 2017.
\newblock Snapshot ensembles: Train 1, get m for free.
\newblock \emph{ICLR} .

\bibitem[{Huang et~al.(2018{\natexlab{b}})Huang, Liu, van~der Maaten, and
  Weinberger}]{huang2018condensenet}
Huang, G.; Liu, S.; van~der Maaten, L.; and Weinberger, K. 2018{\natexlab{b}}.
\newblock {CondenseNet}: An efficient densenet using learned group
  convolutions.
\newblock In \emph{CVPR}.

\bibitem[{Ilg et~al.(2018)Ilg, Cicek, Galesso, Klein, Makansi, Hutter, and
  Brox}]{ilg2018uncertainty}
Ilg, E.; Cicek, O.; Galesso, S.; Klein, A.; Makansi, O.; Hutter, F.; and Brox,
  T. 2018.
\newblock Uncertainty estimates and multi-hypotheses networks for optical flow.
\newblock In \emph{ECCV}.

\bibitem[{Kim et~al.(2017)Kim, Park, Kim, and Hwang}]{kim2017splitnet}
Kim, J.; Park, Y.; Kim, G.; and Hwang, S.~J. 2017.
\newblock SplitNet: Learning to semantically split deep networks for parameter
  reduction and model parallelization.
\newblock In \emph{ICML}.

\bibitem[{Krizhevsky(2009)}]{krizhevsky2009learning}
Krizhevsky, A. 2009.
\newblock \emph{Learning Multiple Layers of Features from Tiny Images}.
\newblock Master's thesis, University of Toronto.

\bibitem[{Krogh and Vedelsby(1995)}]{krogh1995neural}
Krogh, A.; and Vedelsby, J. 1995.
\newblock Neural network ensembles, cross validation, and active learning.
\newblock In \emph{NIPS}.

\bibitem[{Lakshminarayanan, Pritzel, and
  Blundell(2017)}]{lakshminarayanan2017simple}
Lakshminarayanan, B.; Pritzel, A.; and Blundell, C. 2017.
\newblock Simple and scalable predictive uncertainty estimation using deep
  ensembles.
\newblock In \emph{NIPS}.

\bibitem[{Lan, Zhu, and Gong(2018)}]{lan2018knowledge}
Lan, X.; Zhu, X.; and Gong, S. 2018.
\newblock Knowledge distillation by on-the-fly native ensemble.
\newblock In \emph{NIPS}.

\bibitem[{Lee and Chung(2020)}]{lee2019robust}
Lee, J.; and Chung, S.-Y. 2020.
\newblock Robust Training with Ensemble Consensus.
\newblock In \emph{ICLR}.

\bibitem[{Lee et~al.(2015)Lee, Purushwalkam, Cogswell, Crandall, and
  Batra}]{lee2015m}
Lee, S.; Purushwalkam, S.; Cogswell, M.; Crandall, D.; and Batra, D. 2015.
\newblock Why M heads are better than one: Training a diverse ensemble of deep
  networks.
\newblock \emph{arXiv preprint arXiv:1511.06314} .

\bibitem[{Li et~al.(2019)Li, Zhang, Qi, Yang, and Huang}]{li2019improved}
Li, H.; Zhang, H.; Qi, X.; Yang, R.; and Huang, G. 2019.
\newblock Improved Techniques for Training Adaptive Deep Networks.
\newblock In \emph{ICCV}.

\bibitem[{Liu et~al.(2019)Liu, Stehouwer, Jourabloo, and Liu}]{liu2019deep}
Liu, Y.; Stehouwer, J.; Jourabloo, A.; and Liu, X. 2019.
\newblock Deep tree learning for zero-shot face anti-spoofing.
\newblock In \emph{CVPR}.

\bibitem[{Liu et~al.(2017)Liu, Li, Shen, Huang, Yan, and
  Zhang}]{liu2017learning}
Liu, Z.; Li, J.; Shen, Z.; Huang, G.; Yan, S.; and Zhang, C. 2017.
\newblock Learning efficient convolutional networks through network slimming.
\newblock In \emph{ICCV}.

\bibitem[{Loshchilov and Hutter(2017)}]{loshchilov2016sgdr}
Loshchilov, I.; and Hutter, F. 2017.
\newblock {SGDR}: Stochastic gradient descent with warm restarts.
\newblock In \emph{ICLR}.

\bibitem[{Ma et~al.(2018)Ma, Zhang, Zheng, and Sun}]{ma2018shufflenet}
Ma, N.; Zhang, X.; Zheng, H.-T.; and Sun, J. 2018.
\newblock {ShuffleNet V2}: Practical guidelines for efficient {CNN}
  architecture design.
\newblock In \emph{ECCV}.

\bibitem[{Malinin, Mlodozeniec, and Gales(2020)}]{malinin2019ensemble}
Malinin, A.; Mlodozeniec, B.; and Gales, M. 2020.
\newblock Ensemble Distribution Distillation.
\newblock In \emph{ICLR}.

\bibitem[{Minetto, Segundo, and Sarkar(2019)}]{minetto2019hydra}
Minetto, R.; Segundo, M.; and Sarkar, S. 2019.
\newblock Hydra: an ensemble of convolutional neural networks for geospatial
  land classification.
\newblock \emph{IEEE Transactions on Geoscience and Remote Sensing} .

\bibitem[{Naftaly, Intrator, and Horn(1997)}]{naftaly1997optimal}
Naftaly, U.; Intrator, N.; and Horn, D. 1997.
\newblock Optimal ensemble averaging of neural networks.
\newblock \emph{Network: Computation in Neural Systems} .

\bibitem[{Neal et~al.(2018)Neal, Mittal, Baratin, Tantia, Scicluna,
  Lacoste-Julien, and Mitliagkas}]{neal2018modern}
Neal, B.; Mittal, S.; Baratin, A.; Tantia, V.; Scicluna, M.; Lacoste-Julien,
  S.; and Mitliagkas, I. 2018.
\newblock A modern take on the bias-variance tradeoff in neural networks.
\newblock \emph{arXiv preprint arXiv:1810.08591} .

\bibitem[{Rokach(2010)}]{rokach2010ensemble}
Rokach, L. 2010.
\newblock Ensemble-based classifiers.
\newblock \emph{Artificial Intelligence Review} 33(1-2): 1--39.

\bibitem[{Romero et~al.(2015)Romero, Ballas, Kahou, Chassang, Gatta, and
  Bengio}]{romero2014fitnets}
Romero, A.; Ballas, N.; Kahou, S.~E.; Chassang, A.; Gatta, C.; and Bengio, Y.
  2015.
\newblock Fitnets: Hints for thin deep nets.
\newblock \emph{ICLR} .

\bibitem[{Roy, Panda, and Roy(2020)}]{roy2020tree}
Roy, D.; Panda, P.; and Roy, K. 2020.
\newblock {Tree-CNN}: a hierarchical deep convolutional neural network for
  incremental learning.
\newblock \emph{Neural Networks} .

\bibitem[{Ruiz and Verbeek(2019)}]{ruiz2019adaptative}
Ruiz, A.; and Verbeek, J. 2019.
\newblock Adaptative Inference Cost With Convolutional Neural Mixture Models.
\newblock In \emph{ICCV}.

\bibitem[{Russakovsky et~al.(2015)Russakovsky, Deng, Su, Krause, Satheesh, Ma,
  Huang, Karpathy, Khosla, Bernstein, Berg, and Fei-Fei}]{ILSVRC15}
Russakovsky, O.; Deng, J.; Su, H.; Krause, J.; Satheesh, S.; Ma, S.; Huang, Z.;
  Karpathy, A.; Khosla, A.; Bernstein, M.; Berg, A.~C.; and Fei-Fei, L. 2015.
\newblock {ImageNet} Large Scale Visual Recognition Challenge.
\newblock \emph{IJCV} .

\bibitem[{Sandler et~al.(2018)Sandler, Howard, Zhu, Zhmoginov, and
  Chen}]{sandler2018mobilenetv2}
Sandler, M.; Howard, A.; Zhu, M.; Zhmoginov, A.; and Chen, L.-C. 2018.
\newblock {MobileNetV2}: Inverted residuals and linear bottlenecks.
\newblock In \emph{CVPR}.

\bibitem[{Schwartz et~al.(2019)Schwartz, Dodge, Smith, and
  Etzioni}]{schwartz2019green}
Schwartz, R.; Dodge, J.; Smith, N.~A.; and Etzioni, O. 2019.
\newblock Green AI. CoRR abs/1907.10597 (2019).
\newblock \emph{arXiv preprint arXiv:1907.10597} .

\bibitem[{Song and Chai(2018)}]{song2018collaborative}
Song, G.; and Chai, W. 2018.
\newblock Collaborative learning for deep neural networks.
\newblock In \emph{NIPS}.

\bibitem[{Strubell, Ganesh, and McCallum(2019)}]{strubell2019energy}
Strubell, E.; Ganesh, A.; and McCallum, A. 2019.
\newblock Energy and policy considerations for deep learning in NLP.
\newblock \emph{arXiv preprint arXiv:1906.02243} .

\bibitem[{Sun et~al.(2019)Sun, Cheng, Gan, and Liu}]{sun19emnlp}
Sun, S.; Cheng, Y.; Gan, Z.; and Liu, J. 2019.
\newblock Patient Knowledge Distillation for {BERT} Model Compression.
\newblock In \emph{EMNLP}.

\bibitem[{Tan et~al.(2019)Tan, Chen, Pang, Vasudevan, Sandler, Howard, and
  Le}]{tan2019mnasnet}
Tan, M.; Chen, B.; Pang, R.; Vasudevan, V.; Sandler, M.; Howard, A.; and Le, Q.
  2019.
\newblock {MnasNet}: Platform-aware neural architecture search for mobile.
\newblock In \emph{CVPR}.

\bibitem[{Tanno et~al.(2019)Tanno, Arulkumaran, Alexander, Criminisi, and
  Nori}]{tanno2018adaptive}
Tanno, R.; Arulkumaran, K.; Alexander, D.; Criminisi, A.; and Nori, A. 2019.
\newblock Adaptive neural trees.
\newblock \emph{ICML} .

\bibitem[{Ueda and Nakano(1996)}]{ueda1996generalization}
Ueda, N.; and Nakano, R. 1996.
\newblock Generalization error of ensemble estimators.
\newblock In \emph{Proceedings of International Conference on Neural Networks}.
  IEEE.

\bibitem[{Veit and Belongie(2018)}]{veit2018convolutional}
Veit, A.; and Belongie, S. 2018.
\newblock Convolutional networks with adaptive inference graphs.
\newblock In \emph{ECCV}.

\bibitem[{Wang et~al.(2018)Wang, Yu, Dou, Darrell, and
  Gonzalez}]{wang2018skipnet}
Wang, X.; Yu, F.; Dou, Z.-Y.; Darrell, T.; and Gonzalez, J. 2018.
\newblock {SkipNet}: Learning dynamic routing in convolutional networks.
\newblock In \emph{ECCV}.

\bibitem[{Wu et~al.(2019)Wu, Dai, Zhang, Wang, Sun, Wu, Tian, Vajda, Jia, and
  Keutzer}]{wu2019fbnet}
Wu, B.; Dai, X.; Zhang, P.; Wang, Y.; Sun, F.; Wu, Y.; Tian, Y.; Vajda, P.;
  Jia, Y.; and Keutzer, K. 2019.
\newblock {FBNet}: Hardware-aware efficient ConvNet design via differentiable
  neural architecture search.
\newblock In \emph{CVPR}.

\bibitem[{Wu et~al.(2018)Wu, Nagarajan, Kumar, Rennie, Davis, Grauman, and
  Feris}]{wu2018blockdrop}
Wu, Z.; Nagarajan, T.; Kumar, A.; Rennie, S.; Davis, L.; Grauman, K.; and
  Feris, R. 2018.
\newblock {BlockDrop}: Dynamic inference paths in residual networks.
\newblock In \emph{CVPR}.

\bibitem[{Xie et~al.(2017)Xie, Girshick, Doll{\'a}r, Tu, and
  He}]{xie2017aggregated}
Xie, S.; Girshick, R.; Doll{\'a}r, P.; Tu, Z.; and He, K. 2017.
\newblock Aggregated residual transformations for deep neural networks.
\newblock In \emph{CVPR}.

\bibitem[{Yang et~al.(2020)Yang, Han, Chen, Song, Dai, and
  Huang}]{yang2020resolution}
Yang, L.; Han, Y.; Chen, X.; Song, S.; Dai, J.; and Huang, G. 2020.
\newblock Resolution Adaptive Networks for Efficient Inference.
\newblock In \emph{CVPR}.

\bibitem[{Yu and Huang(2019)}]{yu2019universally}
Yu, J.; and Huang, T. 2019.
\newblock Universally slimmable networks and improved training techniques.
\newblock \emph{ICCV} .

\bibitem[{Yu et~al.(2020)Yu, Jin, Liu, Bender, Kindermans, Tan, Huang, Song,
  Pang, and Le}]{yu2020bignas}
Yu, J.; Jin, P.; Liu, H.; Bender, G.; Kindermans, P.-J.; Tan, M.; Huang, T.;
  Song, X.; Pang, R.; and Le, Q. 2020.
\newblock Bignas: Scaling up neural architecture search with big single-stage
  models.
\newblock \emph{ECCV} .

\bibitem[{Yu et~al.(2018)Yu, Yang, Xu, Yang, and Huang}]{yu2018slimmable}
Yu, J.; Yang, L.; Xu, N.; Yang, J.; and Huang, T. 2018.
\newblock Slimmable neural networks.
\newblock \emph{ICLR} .

\bibitem[{Zhang, Ren, and Urtasun(2019)}]{zhang2018graph}
Zhang, C.; Ren, M.; and Urtasun, R. 2019.
\newblock Graph hypernetworks for neural architecture search.
\newblock \emph{ICLR} .

\bibitem[{Zhang et~al.(2018)Zhang, Xiang, Hospedales, and Lu}]{zhang2018deep}
Zhang, Y.; Xiang, T.; Hospedales, T.; and Lu, H. 2018.
\newblock Deep mutual learning.
\newblock In \emph{CVPR}.

\bibitem[{Zhou, Wu, and Tang(2002)}]{zhou2002ensembling}
Zhou, Z.-H.; Wu, J.; and Tang, W. 2002.
\newblock Ensembling neural networks: many could be better than all.
\newblock \emph{Artificial intelligence} .

\end{thebibliography}

\clearpage

\begin{appendices}
\onecolumn

\begin{center}
    \textbf{\Large{---Supplementary Material---}}
\end{center}

\section{HNE Architectures and training details}
\label{ap:architectures}
In the following, we provide a detailed description of our HNE implementation using ResNet \cite{he2016deep} and MobileNetV2 \cite{sandler2018mobilenetv2} architectures for the CIFAR-10/100 and ImageNet datasets, respectively. 

\begin{figure}[h]
  \centering
  \includegraphics[width=0.65\linewidth]{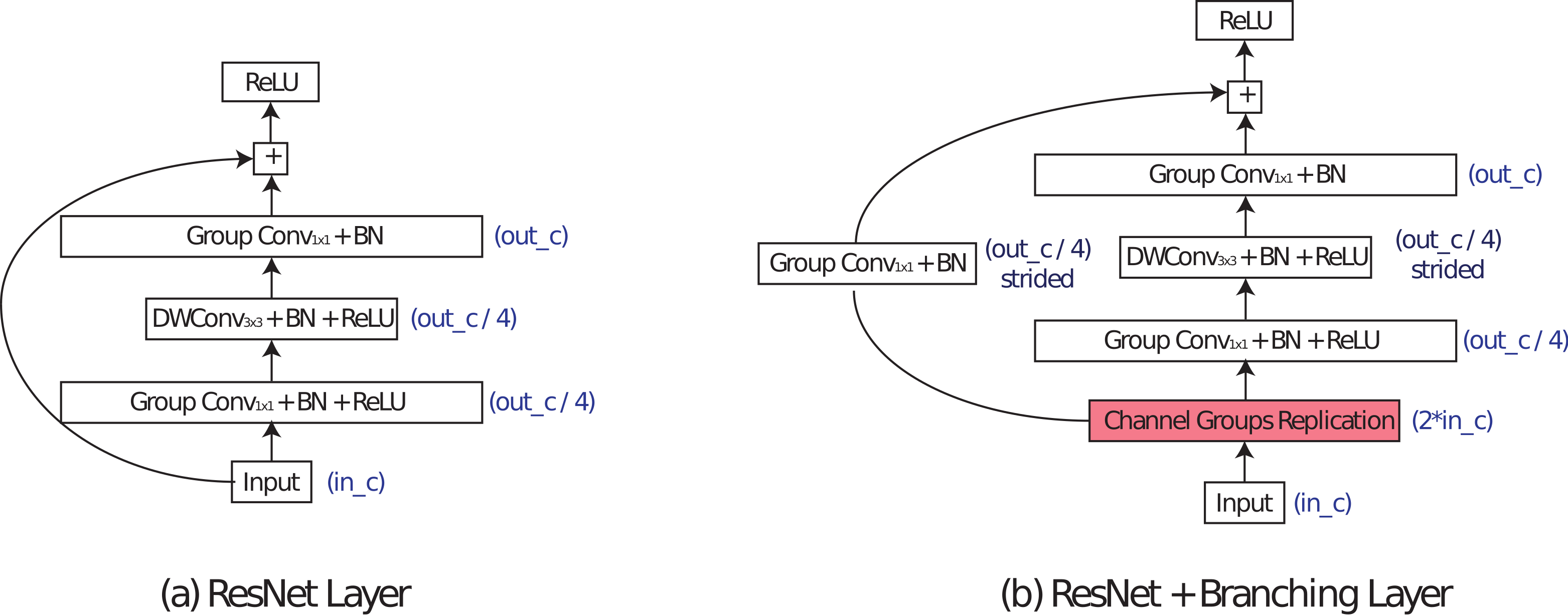}
  \caption{The two  layers used in HNE with ResNet architecture. 
  We use \textsf{in\textunderscore c} and \textsf{out\textunderscore c} to refer to the number of input and output feature channels of the convolutional block as described in Table \ref{tab:HNE_ResNet}. In blue, we show the number of channels resulting after applying each block and whether strided convolution is used or not. BN refers to Batch Normalization.  }
\label{fig:resnet_block}
\end{figure}

\mypar{ResNet HNE} For the CIFAR-10/100 datasets we build our HNE using the two types of layers illustrated in Figure \ref{fig:resnet_block}. We refer to them as ResNet and ResNet+Branching. The first one implements a standard residual block with bottleneck \cite{he2016deep}. 
However, as described in Figure \ref{fig:group_conv} of the main paper, we replace standard convolutional filters by group convolutions in order to compute in parallel the output of the different branches in the tree structure. Additionally, we also replace the standard $3 \times 3$ convolution by a depth-wise separable $3 \times 3$ convolution \cite{howard2017mobilenets} in order to reduce computational complexity. 
The ResNet+Branching layer follows the structure of ResNets ``shortcut'' blocks used when the image resolution is reduced or the number of channels is augmented. In HNE we  employ it when a branch is split in two. Therefore, we add a channel replication operation as shown in Figure \ref{fig:resnet_block}(b). Using these two types of layers, we implement HNE with $B=5$ blocks embedding a total of $16$ ResNets of depth $50$. 
See Table \ref{tab:HNE_ResNet} for a detailed description of the full architecture configuration.

\begin{table*}[h]
\centering
\resizebox{0.95\textwidth}{!}{%
\begin{tabular}{|c|c|c|c|c|c|c|c|c|c|}
\rowcolor[HTML]{FFFFFF} 
\textbf{\begin{tabular}[c]{@{}c@{}}Block.\\ Index\end{tabular}} & \textbf{\begin{tabular}[c]{@{}c@{}}Layer\\ Type\end{tabular}} & \textbf{Stride} & \textbf{Repetitions} & \textbf{\begin{tabular}[c]{@{}c@{}}Input\\ Res.\end{tabular}} & \textbf{\begin{tabular}[c]{@{}c@{}}Output\\ Res.\end{tabular}} & \textbf{\begin{tabular}[c]{@{}c@{}}Input\\ Groups ($G_i$)\end{tabular}} & \textbf{\begin{tabular}[c]{@{}c@{}}Output \\ Groups ($G_o$)\end{tabular}} & \textbf{\begin{tabular}[c]{@{}c@{}}Input\\ Channels\end{tabular}} & \textbf{\begin{tabular}[c]{@{}c@{}}Output\\ Channels\end{tabular}} \\ \hline
\rowcolor[HTML]{EFEFEF} 
0 & Conv$_{3 \times 3} $ + BN + ReLU & 1 & 1 & $32 \times 32$ & $32 \times 32$ & 1 & 1 & 3 & 16 \\
\rowcolor[HTML]{EFEFEF} 
0 & Conv$_{1 \times 1}$  + BN + ReLU & 1 & 1 & $32 \times 32$ & $32 \times 32$ & 1 & 1 & 16 & 64 \\
\rowcolor[HTML]{FFFFFF} 
1 & ResNet+Branching & 1 & 1 & $32 \times 32$ & $32 \times 32$ & 1 & 2 & 64 & 64 $G_o$ \\
\rowcolor[HTML]{FFFFFF} 
1 & ResNet & 1 & 2 & $32 \times 32$ & $32 \times 32$ & 2 & 2 & 64 $G_i$ & 64 $G_o$ \\
\rowcolor[HTML]{EFEFEF} 
2 & ResNet+Branching & 2 & 1 & $32 \times 32$ & $16 \times 16$ & 2 & 4 & 64 $G_i$ & 128 $G_o$ \\
\rowcolor[HTML]{EFEFEF} 
2 & ResNet & 1 & 3 & $16 \times 16$ & $16 \times 16$ & 4 & 4 & 128 $G_i$ & 128 $G_o$ \\
\rowcolor[HTML]{FFFFFF} 
3 & ResNet+Branching & 2 & 1 & $16 \times 16$ & $8 \times 8$ & 4 & 8 & 128 $G_i$ & 256 $G_o$ \\
\rowcolor[HTML]{FFFFFF} 
3 & ResNet & 1 & 3 & $8 \times 8$ & $8 \times 8$ & 8 & 8 & 256 $G_i$ & 256 $G_o$ \\
\rowcolor[HTML]{EFEFEF} 
4 & ResNet+Branching & 2 & 1 & $8 \times 8$ & $8 \times 8$ & 8 & 16 & 256 $G_i$ & 256 $G_o$ \\
\rowcolor[HTML]{EFEFEF} 
4 & ResNet & 1 & 4 & $8 \times 8$ & $8 \times 8$ & 16 & 16 & 256 $G_i$ & 256 $G_o$ \\
\rowcolor[HTML]{FFFFFF} 
{\color[HTML]{000000} Classifier} & {\color[HTML]{000000} Avg. Pool + Linear} & {\color[HTML]{000000} 1} & {\color[HTML]{000000} 1} & {\color[HTML]{000000} $8 \times 8$} & {\color[HTML]{000000} $1\times1$} & {\color[HTML]{000000} 16} & {\color[HTML]{000000} 16} & {\color[HTML]{000000} 256 $G_i$} & {\color[HTML]{000000} $L$ $G_o$}
\end{tabular}%
}
\caption{Full layer configuration of HNE based on ResNet architecture. Each row shows: (1) The block index in the hierarchical structure. (2) The type of layer, whether strided convolutions is used and the number of times that it is stacked. (3) Input and output resolutions of the feature maps. (4) The number of input and output groups used for convolutions, representing the number of active branches in the tree structure. (5) The number of input and output channels before and after applying the layers. Note that it is multiplied by the number of input and output groups representing the number of active branches. Finally, $L$ refers to the number of classes in the dataset.}
\label{tab:HNE_ResNet}
\end{table*}

\begin{figure}[h]
  \centering
  \includegraphics[width=0.65\linewidth]{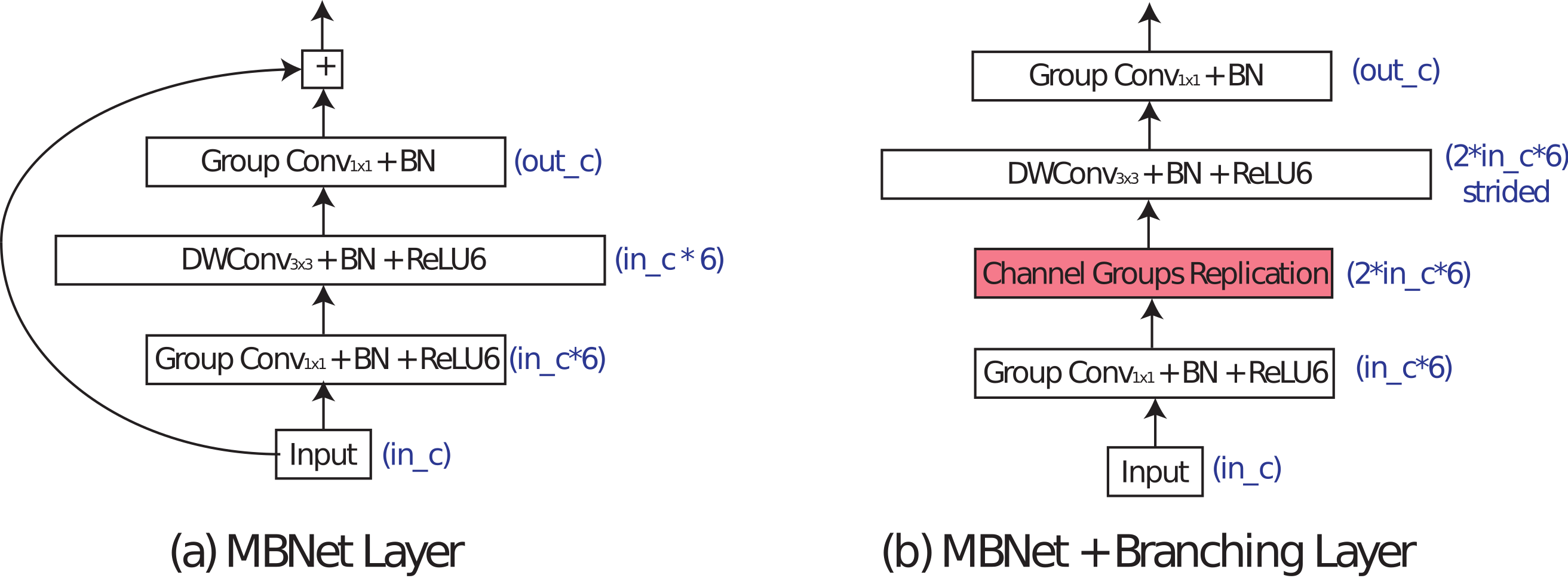}
  \caption{
  The two  layers used in HNE with  MobileNet architectures. 
  We use \textsf{in\textunderscore c} and \textsf{out\textunderscore c} to refer to the number of input and output feature channels of the convolutional block as described in Table \ref{tab:HNE_MBNet}. In blue, we show the number of channels resulting after applying each block and whether strided convolution is used. BN refers to Batch Normalization. }
    \label{fig:mbnet_block}
\end{figure}

\mypar{MobileNet HNE} In order to implement an efficient ensemble of MobileNetV2 networks, we employ the same layers used in the original architecture (see Figure \ref{fig:mbnet_block}). In this case, we refer to them as MBNet and MBNet+Branching layers. The first one implements the inverted residual blocks used in MobileNetV2 but using group convolutions to account for the different HNE branches. The second one is analogous to the shortcut blocks used in this model but we add a channel group replication to implement the branch splits in the tree structure. The detailed layer configuration is shown in Table \ref{tab:HNE_MBNet}. Note that our MobileNet HNE embeds a total of $16$ networks with the same architecture as a MobileNetV2 model with channel width multiplier of $1.4$. The only difference is that we replace the last convolutional layer by a $1 \times 1$ convolution and a fully-connected layer, as suggested in \cite{howard2019searching}. This modification offers a reduction of the total number of FLOPs without a significant drop in accuracy.

\begin{table}[h]
\centering
\resizebox{0.95\textwidth}{!}{%
\begin{tabular}{|c|c|c|c|c|c|c|c|c|c}
\rowcolor[HTML]{FFFFFF} 
\textbf{\begin{tabular}[c]{@{}c@{}}Block.\\ Index\end{tabular}} & \textbf{\begin{tabular}[c]{@{}c@{}}Layer\\ Type\end{tabular}} & \textbf{Stride} & \textbf{Repetitions} & \textbf{\begin{tabular}[c]{@{}c@{}}Input\\ Res.\end{tabular}} & \textbf{\begin{tabular}[c]{@{}c@{}}Output\\ Res.\end{tabular}} & \textbf{\begin{tabular}[c]{@{}c@{}}Input\\ Groups ($G_i$)\end{tabular}} & \textbf{\begin{tabular}[c]{@{}c@{}}Output \\ Groups ($G_o$)\end{tabular}} & \textbf{\begin{tabular}[c]{@{}c@{}}Input\\ Channels\end{tabular}} & \multicolumn{1}{c|}{\cellcolor[HTML]{FFFFFF}\textbf{\begin{tabular}[c]{@{}c@{}}Output\\ Channels\end{tabular}}} \\ \hline
\rowcolor[HTML]{EFEFEF} 
0 & Conv$_{3 \times 3} $ + BN + ReLU6 & 2 & 1 & $224 \times 224$ & $112 \times 112$ & 1 & 1 & 3 & 24 \\
\rowcolor[HTML]{EFEFEF} 
0 & DWConv$_{3 \times 3} $ + BN + ReLU6 & 1 & 1 & $112 \times 112$ & $112 \times 112$ & 1 & 1 & 24 & 24 \\
\rowcolor[HTML]{EFEFEF} 
0 & MBNet & 2 & 2 & $112 \times 112$ & $56 \times 56$ & 1 & 1 & 24 & 32 \\
\rowcolor[HTML]{FFFFFF} 
1 & MBNet+Branching & 2 & 1 & $56 \times 56$ & $28 \times 28$ & 1 & 2 & 32 & \multicolumn{1}{c|}{\cellcolor[HTML]{FFFFFF}{\color[HTML]{330001} 48 $  G_o$}} \\
\rowcolor[HTML]{FFFFFF} 
1 & MBNet & 1 & 2 & $28 \times 28$ & $28 \times 28$ & 2 & 2 & 48 $  G_i$ & \multicolumn{1}{c|}{\cellcolor[HTML]{FFFFFF}48 $  G_o$} \\
\rowcolor[HTML]{EFEFEF} 
2 & MBNet+Branching & 2 & 1 & $28 \times 28$ & $14 \times 14$ & 2 & 4 & 48 $  G_i$ & \multicolumn{1}{c|}{\cellcolor[HTML]{EFEFEF}88 $  G_o$} \\
\rowcolor[HTML]{EFEFEF} 
2 & MBNet & 1 & 3 & $14 \times 14$ & $14 \times 14$ & 4 & 4 & 88 $  G_i$ & \multicolumn{1}{c|}{\cellcolor[HTML]{EFEFEF}88 $  G_o$} \\
\rowcolor[HTML]{FFFFFF} 
3 & MBNet+Branching & 2 & 1 & $28 \times 28$ & $14 \times 14$ & 4 & 8 & 88 $  G_i$ & \multicolumn{1}{c|}{\cellcolor[HTML]{FFFFFF}136 $  G_o$} \\
\rowcolor[HTML]{FFFFFF} 
3 & MBNet & 1 & 2 & $14 \times 14$ & $14 \times 14$ & 8 & 8 & 136 $  G_i$ & \multicolumn{1}{c|}{\cellcolor[HTML]{FFFFFF}136 $  G_o$} \\
\rowcolor[HTML]{EFEFEF} 
4 & MBNet+Branching & 2 & 1 & $14 \times 14$ & $7 \times 7$ & 8 & 16 & 136 $  G_i$ & \multicolumn{1}{c|}{\cellcolor[HTML]{EFEFEF}224 $  G_o$} \\
\rowcolor[HTML]{EFEFEF} 
4 & MBNet & 1 & 2 & $7 \times 7$ & $7 \times 7$ & 16 & 16 & 224 $  G_i$ & \multicolumn{1}{c|}{\cellcolor[HTML]{EFEFEF}224 $  G_o$} \\
\rowcolor[HTML]{EFEFEF} 
4 & Conv$_{1 \times 1} $ + BN + ReLU6 & 1 & 1 & $7 \times 7$ & $7 \times 7$ & 16 & 16 & 224 $  G_i$ & \multicolumn{1}{c|}{\cellcolor[HTML]{EFEFEF}1344 $  G_o$} \\
\rowcolor[HTML]{EFEFEF} 
4 & Avg. Pool + Linear  + ReLU6 & 1 & 1 & $7 \times 7$ & $1 \times 1$ & 16 & 16 & 1344 $  G_i$ & \multicolumn{1}{c|}{\cellcolor[HTML]{EFEFEF}1792 $  G_o$} \\
\rowcolor[HTML]{FFFFFF} 
Classifier &  Linear & 1 & 1 & $1 \times 1$ & $1\times1$ & 16 & 16 & 1792 $  G_i$ & \multicolumn{1}{c|}{\cellcolor[HTML]{FFFFFF}1000 $  G_o$} 
\end{tabular}%
}
\caption{Full layer configuration of HNE based on MobileNetV2 architecture. 
Each row shows: (1) The block index in the hierarchical structure. (2) The type of layer, whether strided convolutions is used and the number of times that it is stacked. (3) Input and output resolutions of the feature maps. (4) The number of input and output groups used for convolutions, representing the number of active branches in the tree structure. (5) The number of input and output channels before and after applying the layers. Note that it is multiplied by the number of input and output groups representing the number of active branches. 
}
\label{tab:HNE_MBNet}
\end{table}

\mypar{Hyper-parameter settings} 
All the networks are optimized  using SGD with initial learning rate of $0.1$ and a momentum of $0.9$. For CIFAR datasets, following \cite{ruiz2019adaptative}, we train HNE using a cosine-shaped learning rate scheduler. However, for faster experimentation we reduce the number of epochs to $200$ and use a batch size of $128$. Weight decay is set to $5\mathrm{e}{-4}$. 
For ImageNet, we use the same training hyper-parameters but using a total of $120$ epochs with a batch size of $256$ split accross 8 GPUs. Weight decay is set to $1\mathrm{e}{-5}$. For distillation methods, following \cite{song2018collaborative}, we use a temperature $T=2$ to generate soft-labels from ensemble predictions. 
Finally, we set the trade-off parameter between the cross-entropy loss and the distillation loss to $\alpha=0.5$, unless  specified otherwise. All  experiments were conducted on a cluster of NVIDIA GPUs.

\clearpage
\section{Complementary Results on Hierarchical Distillation}
\begin{figure*}[ht]
\centering
\includegraphics[width=0.95\linewidth]{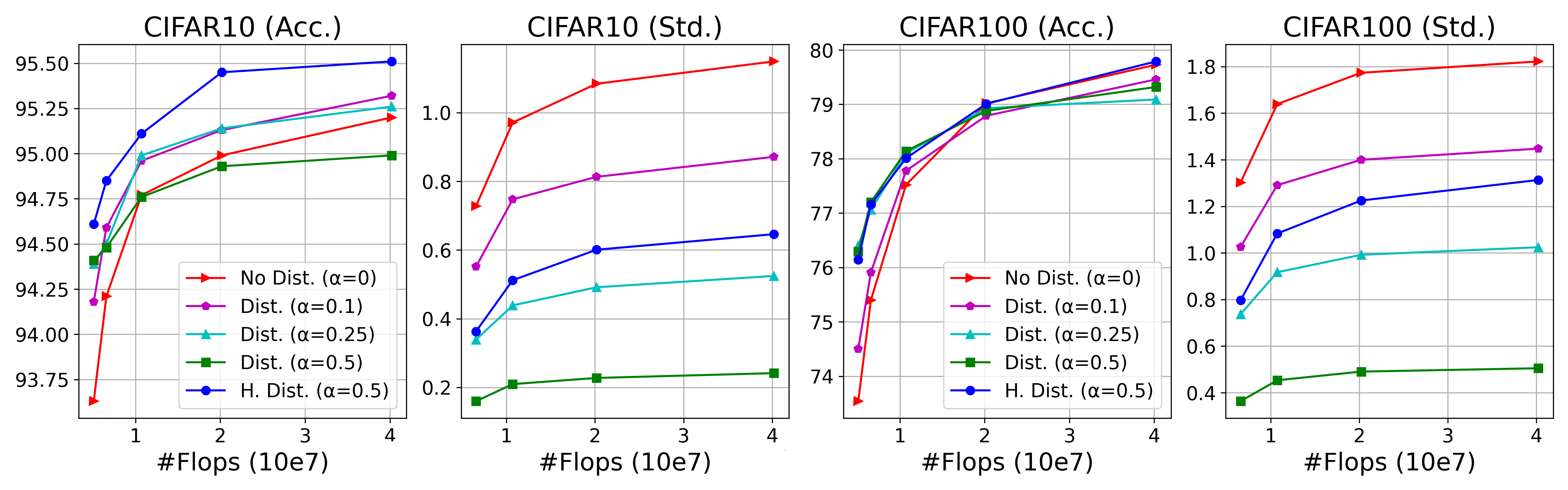}
\caption{
Accuracy and standard deviation in logits \textit{vs} FLOPs  for HNE trained (i) without  distillation, (ii) with  co-distillation,  and (iii) with our hierarchical distillation. Curves represent the evaluation of ensembles of size 1 up to 16.}
\label{fig:distillation_experiment2}
\figvspace
\end{figure*}

We provide the results for the bigger HNE model corresponding to the experiment shown in Fig. \ref{fig:distillation_experiment2}. From the reported results, we can extract similar conclusions as the ones discussed in the main paper for HNE$_{small}$. 

\clearpage

\end{appendices}
\end{document}